\RequirePackage{fix-cm}
\documentclass[twocolumn]{svjour3}          
\smartqed  
\usepackage{graphicx}
\usepackage{amsmath}
\usepackage{multirow}
\usepackage{pifont}
\usepackage{graphicx}
\usepackage{amsmath}
\usepackage{amssymb}
\usepackage{array}
\usepackage{makecell, multirow, diagbox, nicefrac}
\usepackage{pifont}
\usepackage{natbib}
\usepackage[title]{appendix}%

\usepackage[dvipsnames, svgnames, x11names, table]{xcolor}
\definecolor{green}{rgb}{1,0,0}

\usepackage{enumerate}
\usepackage{graphicx}
\usepackage{svg}
\usepackage{tablefootnote}
\usepackage{caption}
\usepackage{subcaption}
\usepackage{color}
\usepackage{algorithm}
\usepackage{algorithmic}
\usepackage[algo2e]{algorithm2e}
\usepackage{makecell, multirow, diagbox}
\usepackage{extarrows}
\usepackage{makecell}
\usepackage{booktabs}
\usepackage{amssymb}
\usepackage{bbding}
\usepackage{amsmath}

\usepackage{xcolor}

\usepackage{xspace}
\usepackage[
    colorlinks,
    linkcolor=blue,
    anchorcolor=blue,
    citecolor=blue,
    CJKbookmarks=True
]{hyperref}

\begin{document}

\title{Fast-ParC: Capturing Position Aware Global Feature for ConvNets and ViTs
\thanks{Work done during T. Yang was an intern at Intellifusion. A preliminary version of this work has been presented in the ECCV 2022~\citep{ParCNet}. H. Zhang is the corresponding author.
}
}

\author{Tao Yang,
 Haokui Zhang,
 Wenze Hu,
 Changwen Chen,
 Xiaoyu Wang 
}


\institute{
           Tao Yang, Changwen Chen\at
           $^1$Hong Kong Polytechnic University, China
           \and
           Haohkui Zhang\at
           $^2$Northwestern Polytechnical University, China
           \and
           Haokui Zhang, Wenze Hu\at
           $^3$Intellifusion, China
           \and
           Xiaoyu Wang\at
           $^4$Hong Kong University of Science and Technology  (Guangzhou), China
}
\date{Received: date / Accepted: date}

\maketitle

\begin{abstract}

Transformer models have made tremendous progress in various fields in recent years. In computer vision, ViTs (vision transformers) also become strong alternatives to ConvNets (convolutional neural networks), yet they have not been able to replace ConvNets since both have their distinct merits. For instance, ViTs are good at extracting global features with attention mechanisms, while ConvNets are more efficient in modeling local relationships due to their strong inductive bias. One natural idea is to combine the strengths of both ConvNets and ViTs to design new structures. 
This paper proposes a new basic neural network operation named position-aware circular convolution (ParC) and its accelerated version Fast-ParC. The ParC operation can capture global features using a global kernel (i.e., with a kernel size identical to feature resolution) and circular convolution while keeping location sensitiveness by employing position embedding. Compared with MHA (Multi-Head Attention), the ParC effectively reduces the time complexity of global operation from $\mathcal{O}(n^2)$ to $\mathcal{O}(n^{3/2})$. Leveraging the FFT (Fast Fourier Transform), Fast-ParC further reduces this complexity of ParC to $\mathcal{O}(n\log n)$. The proposed operation can be used in a plug-and-play manner to 1) convert ViTs to pure-ConvNet architecture to enjoy broader hardware support and achieve higher inference speed; 2) replace traditional convolutions in the deep stage of ConvNets to improve accuracy by enlarging the effective receptive field. Experiment results show that our ParC op can effectively enlarge the receptive field of traditional ConvNets, and adopting the proposed operation benefits both ViTs and ConvNet models on all three popular vision tasks: image classification, object detection, and semantic segmentation. Source code will be available at \url{https://github.com/yangtao2019yt/Fast_ParC.git}.


\end{abstract}

\begin{figure*}[t]
    \centering
    \includegraphics[scale=0.44]{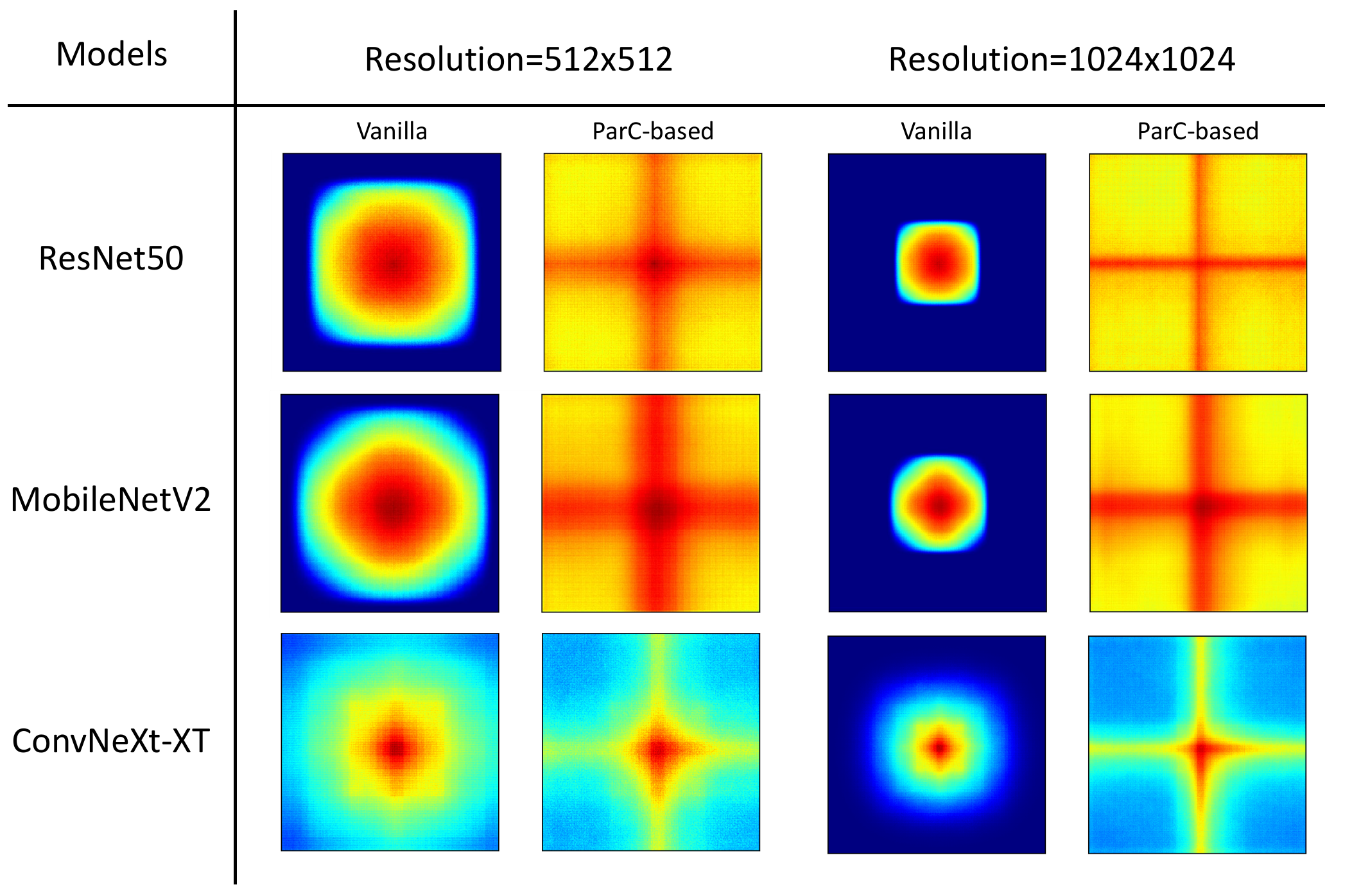}
    \caption{Visualization of the Effective Receptive Field(ERF)~\citep{ERF2017} of different convolutional neural networks(CNNs). ParC significantly increases the CNNs' ERF to global. Code is based on the repository published by RepLKNet~\citep{RepKLNet}.}
    \label{fig:parc_erf_visualization}
\end{figure*}

\section{Introduction}
\label{sec:introduction}
Vision transformer has been a rising star recently. Transformer was first proposed in 2017 to solve the NLP task~\citep{attention}. In 2020, \cite{ViT} directly applied the original transformer to the image classification task and found it achieved better results than convolutional networks (ConvNets) when pre-trained on large datasets~\cite[e.g., JFM-300M][]{JFT_300M}. ViT and its variants~\citep{Swin, PVT} are then widely used on other downstream vision tasks such as object detection~\citep{DETR, YOLOS}, semantic segmentation~\citep{SETR}, and multi-modal tasks like human-object interactions (HOI)~\citep{UPT}, text-to-image (T2I) synthesis~\citep{DALLE2}, etc. Despite the great success of transformers, they still cannot replace ConvNets completely. As summarized in previous works \citep{ParCNet, CMT, MobileViT, MobileFormer}, when compared with ViTs, ConvNets are easier to train and enjoy superior hardware support. In addition, ConvNets still dominate in the field of lightweight models~\citep{MobileNetV2, shufflenetv2, efficientnet} designed for mobile devices, and other edge computing scenarios.

Both transformers and ConvNets have unique characteristics. Transformers, known for their MHA (Multi-Head Attention) mechanism, are adept at capturing long-range pairwise relationships between tokens, granting them superior global modeling prowess. However, this comes at the cost of increased computational demands. The time complexity of MHA grows quadratically with the number of tokens, making it inefficient to process high-resolution feature maps. Contrastingly, convolution operations excel at gleaning local information with a compact sliding window (typically 3$\times$3 in size). The convolution could be considered as an efficient matrix multiplication with a unique weight-reusing scheme, in which the same convolution kernel is reused across varied input data and spatial positions, ensuring the linear growth in FLOPs with respect to the feature resolution. 
Furthermore, given its longer-standing presence in the field, ConvNets has accrued other unique benefits, most notably in hardware support. A variety of acceleration schemes exist for the hardware implementations of convolution, such as Winograd~\citep{winograd}, FFT~\citep{FFT-NVIDIA}, im2col~\citep{im2col}), which are applicable across a range of platforms, from general-purpose ones like CPUs and GPUs to specialized accelerators like FPGAs and ASICs. In essence, while convolution operations are more cost-effective to implement, they lack the ability to capture global relationships in the way MHA can. It is clear that a complementary relationship exists between the representational capabilities of transformers and the efficiency of ConvNets, both of which are indispensable for practical applications.

There are some recent works that combine the merits of transformers and ConvNets. PVT~\citep{PVT}, Swin~\citep{Swin}, and CoAtNet~\citep{CoAtNet} attempt to reintroduce the inductive bias of convolution (such as its sliding-window strategy) to help transformer models to learn better. Works like LeViT \citep{LeViT}, MobileViT~\citep{MobileViT}, and EfficientFormer~\citep{EfficientFormer} focus on designing efficient hybrid architectures. Most of these works bring the two kinds of networks together. Still, they fail to tackle the critical problem: the extra computational and engineering complexity of the newly introduced attention operation. It's natural to ask: is it possible to design a new operation, which is different from both yet retains the advantage of each?

In this paper, we construct a novel plug-and-play operation named ParC, amalgamating the merits of both transformers and ConvNets. ParC possesses a global receptive field, achieved through the use of global kernels ($K_{h}=H$ or $K_{w}=W$) and the adoption of a circular convolution scheme, both of which two are indispensable. To maintain the positional sensitivity of our operation, we apply an explicit learnable positional embedding in front of the convolution. As depicted in Fig.~\ref{fig:parc_erf_visualization}, the ERFs (Effective Receptive Fields) of various ConvNets can be expanded to global by simply substituting the standard convolution with our proposed ParC operation. As ParC employs pure convolution operation only, it is efficient to be deployed across diverse platforms. To further improve its efficiency, we decompose 2D convolution into two 1D convolutions to mitigate the FLOPs/parameters growth brought by kernel size growth. Based on the above design, we achieve the goal of extracting global features while maintaining low computational complexity. Through experiments, we validate the effectiveness of the new operation in a broad range of tasks and models. In short, the contributions of this paper can be summarized as the following three key points:
\begin{enumerate}
    \item An effective new operation ParC is proposed, combining the merits of both ViTs and ConvNets. Experiments demonstrated the advantages of ParC by applying it to a wide range of models, including MobileViT \citep{MobileViT}, 
    ResNet50 \citep{ResNet}, MobileNetV2 \citep{MobileNetV2} and ConvNeXt \citep{ConvNext}. We also evaluate these models on multiple tasks, including classification, detection, and segmentation. A comparison with dilated convolution is also included to demonstrate its effectiveness.
    \item Fast-ParC is introduced to tackle the issue of prohibitive complexity when applying ParC to high-resolution input features. The Fast-ParC is theoretically equivalent to ParC but significantly more efficient when dealing with feature maps of large resolution (e.g., 112$\times$122). Fast-ParC extends the usage scenarios of ParC, making it a more universally applicable operation.
    \item The internal mechanism of the new operation is analyzed. Through visualization, we elucidate several distinct differences between ParC and vanilla convolution. We show that the Effective Receptive Fields (ERFs) \citep{ERF2017} of vanilla ConvNets are quite limited, whereas the ParC-based networks possess global ERFs indeed. We also demonstrate by Grad-CAM \citep{GradCAM} that ParC-based networks exhibit a more comprehensive focus on the critical regions of images compared to standard ConvNets. 
\end{enumerate}

\section{Related work}
\subsection{Theoretical/Effective Receptive Field}
\cite{cat_RF} found in neuroscience that the neurons in the shallow layer extract local features only, and the scope covered is accumulated layer by layer, called the "receptive field (RF)". Since the success of VGGNet \citep{VGG}, the design of CNN architecture follows a similar pattern \citep{ResNet, DenseNet, MobileNetV2, efficientnet} - using a stacking of small kernels like 3$\times$3 instead of larger kernels. Some previous work gives the theoretical computation of CNN's receptive field \citep{computing_RF, RF_ERF_PRF}, namely theoretical receptive field (TRF) - under which concept, the receptive field of two layers of 3$\times$3 equals one layer of 5$\times$5. Nevertheless, some works \citep{ERF2017, RF_ERF_PRF} cast doubt on this view since, in fact, the importance of pixel degraded quickly from the center to the edge in a feature map. Later, the effective receptive field (ERF) was proposed to measure the region of the input image, which could actually impact the neurons' activation pattern. \cite{ERF2017} back-propagate the center pixel and compute the partial derivative of the input image to examine this region. By studying a sequence of convolution networks, they found the effective receptive field is usually much smaller than their theoretical counterparts. SKNet \citep{SKNet} adopts attention mechanisms in selecting appropriate receptive fields. RF-Next \citep{RFNext} proposes a NAS-based workflow to search the receptive fields for models automatically. These works show that a proper decision of receptive field could be quite beneficial for networks' performance. Recent work also found that enlarging the receptive field of convolution networks can lead to better model performance. We call them "Large Kernel Convolution Network", which will be discussed later in Section~\ref{sec::largeKernel}. 

\subsection{Vision Transformer and Hybrid Structures}
ViTs achieve impressive performance on various vision tasks. However, the original ViT \citep{ViT} has some restrictions. For instance, it is heavyweight, has low computational efficiency, and is hard to train. Subsequent variants of ViTs are proposed to overcome these problems. From the point of improving training strategy, \cite{DeiT} proposed to use knowledge distillation to train ViT models and achieved competitive accuracy with less pre-training data. To further improve the model architecture, some researchers attempted to optimize ViTs by learning from ConvNets. Among them, PVT \citep{PVT} and CVT \citep{CVT} insert convolutional operations into each stage of ViT to reduce the number of tokens and build hierarchical multi-stage structures. Swin transformer \citep{Swin} computes self-attention within shifted local windows. PiT \citep{PiT} jointly uses the pooling layer and depth-wise convolution layer to achieve channel multiplication and spatial reduction. CCNet \citep{CCNet} proposes a simplified version of the self-attention mechanism called criss-cross attention and inserts it into ConvNets to build ConvNets with a global receptive field. These papers clearly show that some techniques of ConvNets can be applied to vision transformers to design better vision transformer models.

Another popular line of research is combining elements of ViTs and ConvNets to design new backbones. Graham et al. mixed ConvNet and transformer in their LeViT~\citep{LeViT} model, which significantly outperforms previous ConvNet and ViT models with respect to the speed/accuracy trade-off. BoTNet~\citep{BoTNet} replaces the standard convolution with multi-head attention in the last few blocks of ResNet. ViT-C~\citep{VIT-C} adds early convolutional stems to vanilla ViT. ConViT~\citep{ConViT} incorporates soft convolutional inductive biases via a gated positional self-attention. The CMT~\citep{CMT} block consists of a depthwise convolution-based local perception unit and a lightweight transformer module. CoatNet~\citep{CoAtNet} merges convolution and self-attention to design a new transformer module, which focuses on both local and global information. 

\subsection{Large Kernel Convolution Network}\label{sec::largeKernel}
Early ConvNets such as AlexNet~\citep{AlexNet} and  GoogleNet~\citep{GoogleNet} uses big kernel like 5$\times$5 or 7$\times$7. But since the success of VGGNet~\citep{VGG}, stacking small kernels like 3$\times$3 and 1$\times$1 becomes believed to be an efficient choice for computation and storage. Recently, inspired by the success of vision transformers, big kernels have been reused as a powerful tool for improving the model performance again. ConvNext~\citep{ConvNext} modernizes a standard ResNet towards the design of a vision transformer by introducing a series of incremental but effective designs, where 7$\times$7 depth-wise convolution is used following the spirit of windowed-SA in Swin~\citep{Swin}. RepLKNet~\citep{RepKLNet} scales up the convolution kernel to 31$\times$31 and obtains a performance gain, but the re-parameterization trick used would burden the training process, and an extra conversion step is needed for model deployment. Later, Rao et al. use an even larger kernel of 51$\times$51 with dynamic sparsity~\citep{51x51}.  GFNet~\citep{GFNet} replaces the SA (self-attention) in transformer blocks with a global Fourier convolution implemented with FFT. 

Our work is most closely related to RepLKNet~\citep{RepKLNet} and GFNet~\citep{GFNet}. Both these methods and our proposed ParC focus on enlarging the effective receptive field, but our proposed op is different from the following perspectives:
1) Our ParC uses learnable position embedding to keep the result feature map position sensitive. This is important for location-sensitive tasks such as semantic segmentation and object detection. Experiments in ablation studies also verify this point. 2) Our ParC adopts lightweight designs. RepLKNet applies large two-dimensional convolution kernels up to 31$\times$31, and GFNet adopts a learnable complex weight matrix with shape $2CHW$, while our ParC operation only needs two one-dimensional convolutions, reducing the parameters to $\frac{1}{2}C(H+W)$.  3) Different from RepLKNet and GFNet, which emphasize network designs holistically, our proposed ParC is a new basic operator that can be inserted into ViTs and ConvNets in a plug-and-play manner. Our experimental results in Section~\ref{subsection:experiment_vit} and \ref{subsection:experiment_convnet} verify this point. In addition, we also propose Fast-ParC, which further broadens the usage scenario of ParC. 

\begin{figure*}[t]
\centering
\includegraphics[scale=0.55]{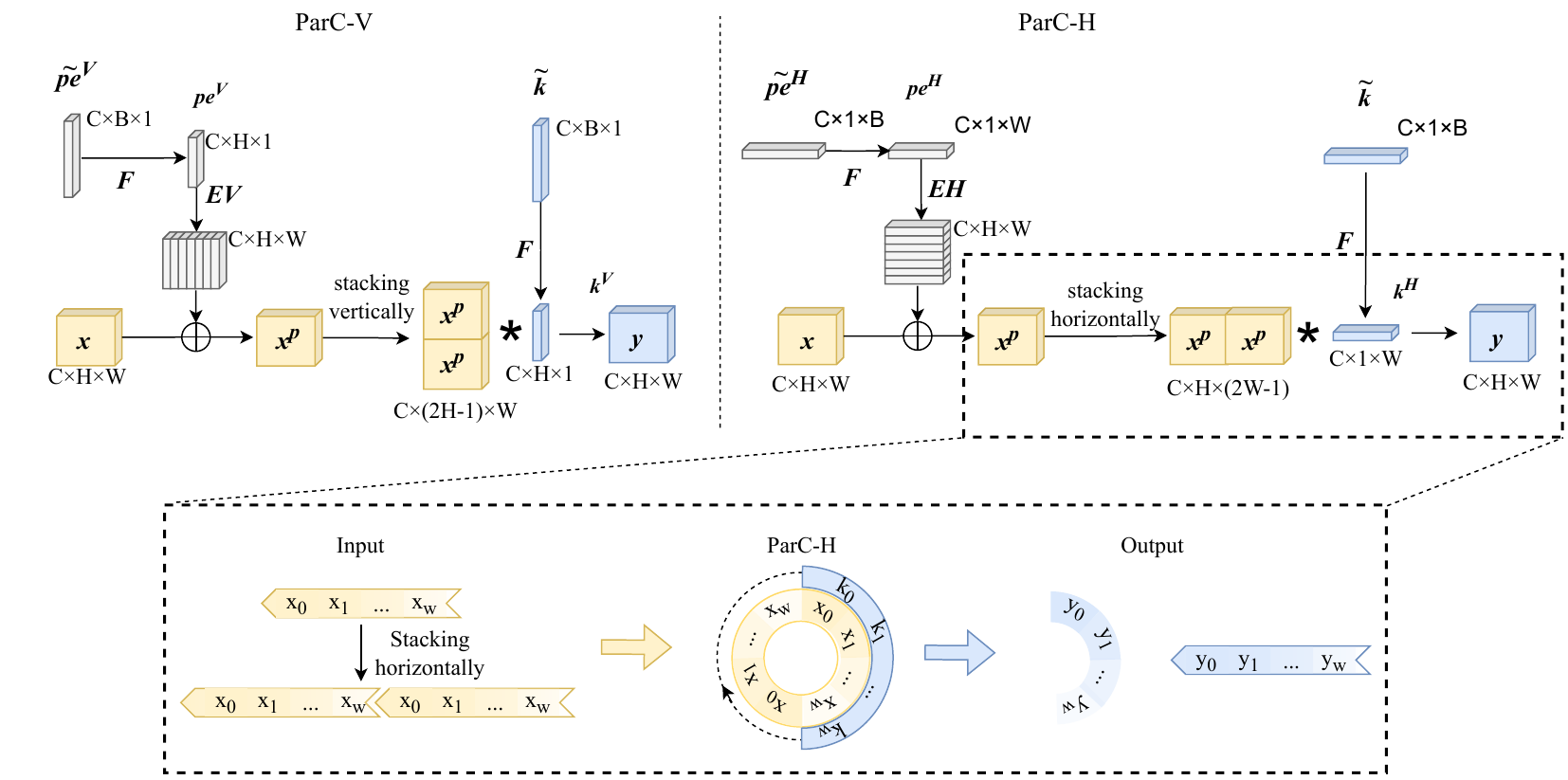}
    \caption{Illustration of the position-aware circular convolution, corresponding with Eq.~\ref{sptail_parc} The circular convolution part is amplified with a dash line connection. ParC-H and ParC-V follow similar strategies but are conducted on different axes.}
\label{fig:parc_operation}
\end{figure*}

\section{The Proposed ParC/Fast-ParC Operation}
In this section, we first introduce the proposed ParC operation by comparing it with the vanilla convolution operation. As a supplement for inputs of large resolutions, the FFT accelerated version of ParC, named Fast-ParC, is then presented. ParC/Fast-ParC are different implementations of the same operation, enjoying strict theoretical equivalence. Finally, we explain how to use the proposed ParC/Fast-ParC in ViT and ConvNet models.   

\subsection{ParC Operation} \label{subsection:parc_operation}
\subsubsection{Vanilla Depth-wise Convolution} \label{subsubsection:vanilla_conv}
To describe a 1D depth-wise convolution conducted in the horizontal dimension (noted as Conv1d-H) on a 4D input tensor shaped as $B\!\times\!C\!\times\!H\!\times\!W$, we could first focus on one specific channel. We denote the output as $\mathbf{y}=\{y_{1},...,y_{H-1}\}$, input as $\mathbf{x}=\{x_{0},x_{1},...,x_{H-1}\}$, the convolution weight as $\mathbf{w}=\{w_{0},w_{1},...,w_{K_{h}-1}\}$. The Pytorch-style convolution (i.e., F.conv1d)  with zero-padding can then be expressed as:
\begin{align} \label{torch_conv}
    y_{i}=\sum_{k=0}^{K_{h}-1}w_{k}\cdot x_{k+i-K_{h}/2},
    \quad
    i=0,1,...,H_{y}-1
\end{align}
where $K_{h}/2$ is used to offset the extra padding of $K_{h}/2$ scalars on both sides of the input, Eq. \ref{torch_conv} shows that $y_{i}$ is a function of its local neighboring input (i.e., $x_{i-K_{h}/2}$, ..., $x_{K_{h}/2-1+i}$), and the size of the neighborhood is controlled by the kernel size $K_{h}$. Consequently, a single layer of small kernel convolution cannot collect long-distance information. To solve this shortcoming of vanilla convolution, we propose our ParC with a global receptive field. 

\subsubsection{ParC: Positional-Aware Circular Convolution}  \label{subsubsection:parc_operation}
Define $\mathbf{w}=\{w_{0},w_{1},...,w_{K_{h}-1}\}$ as the kernel weight, and $\mathbf{pe}=\{pe_{0},pe_{1},...,pe_{K_{h}-1}\}$ as the positional embedding. Corresponding with Fig.~\ref{fig:parc_operation}, ParC could be described as: 
\begin{equation} \label{sptail_parc}
    \begin{aligned}
        & y_{i} = \sum_{k=0}^{H-1}w_{k}^{H}\cdot(x_{(k+i){\rm mod}H}^{p}) \\
        & \mathbf{w}^{H} = f(\mathbf{w}, H) \\
        & \mathbf{x}^{p} = \mathbf{x} + f(\mathbf{pe}, H)
    \end{aligned}
\end{equation}
where $i=0,1,...,H-1$. $\mathbf{w}$ represents the learnable kernel, and $\mathbf{pe}$ denotes position embedding. Here, we adopt the interpolation function $f(\cdot, N)$ (e.g., bi-linear, bi-cubic) to adapt the sizes of kernels and position embedding to the size of input features. Mod denotes the modulo operation.


Compared to vanilla convolution, the ParC operation has four significant differences: 1) global kernel, 2) circular convolution, 3) positional embedding, and 4) 1D decomposition. These designs are essential to extract global features effectively, which is demonstrated later by ablation experiments in Section~\ref{section:ablation_study}. In the following, we will elaborate more on the reasons for these design differences:

\textbf{Global kernel and Circular Convolution.} To capture global relationships throughout the entire input map, ParC employs global kernels. The sizes of these kernels match the size of their corresponding feature maps, represented as $K_{y}=H$ or $K_{x}=W$. In most ConvNets' architectures, this size is halved at each stage. For instance, in ResNet50 or ConvNeXt, the global kernel sizes for the four stages are $[56, 28, 14, 7]$. But simply enlarging the kernel size of the vanilla convolution cannot effectively extract the global relationships. Due to zero padding, even if the kernel size matches the resolution, most kernel weights align with the zero padding, providing limited information beyond the absolute location. For a 2D convolution, this limitation is most pronounced when the kernel aligns with an image's edge, where $3/4$ of the input pixels become zeros. To address this, we recommend to use the circular convolution. With circular convolution, kernel weights consistently align with valid pixels during window sliding, as illustrated in Fig.~\ref{fig:parc_operation}.

\textbf{Positional Embedding.} 
As concluded in prior research~\citep{PEG}, standard convolution can encode positional information when utilizing zero-padding. In contrast, circular convolution periodically reuses the input image, leading to a loss of some positional data. To conquer this problem, we incorporate the learnable position encoding inserted before the circular convolution layer. Subsequent experiments underscore its significance for model performance, particularly for downstream tasks that highly depend on spatial information.

\textbf{1D Decomposition. } 
Lastly, we decompose the 2D convolution and position encoding into $H$(horizontal) and $V$(vertical) components to maintain a manageable model size and computational cost. This approach reduces the parameters and FLOPs from $\mathcal{O}(HW)$ to $\mathcal{O}(H+W)$, offering significant compression, especially for higher resolutions.

\textbf{Implementation of Circular Convolution.} Conceptually, the circular convolution needs to be implemented separately from ordinary convolutions because of the extra modulus operation. Though this operation is not supported directly by F.conv2d, it could be easily implemented by padding the input feature map with its copy using the concatenate function before the calling of ordinary 1D convolution routines (See Algorithm~\ref{algorithm_fast_parc}).


When taking vertical dimension $W$ and the channel dimension $C$ into consideration, the Eq.~\ref{sptail_parc} should be extended as
\begin{align} \label{sptail_parc_full}
    \mathbf{Y}_{i,j,c} = \sum_{k=0}^{H-1}\mathbf{W}_{k,c}^{H}\cdot  (\mathbf{X}_{(k+i){\rm mod}H,j,c}^{p})
\end{align}
with $\forall i\in[0,H\!-\!1]$, $\forall j\in[0,W\!-\!1]$ and $\forall c\in[0,C\!-\!1]$. This is the complete representation of one layer depth-wise ParC-H with channel dimension and feature resolution $H\!\times\!W$. In ResNet50-ParC, we also extend the depth-wise separated ParC to a depth-wise dense version, reintroducing the channel interaction, which can be expressed as:
\begin{align}
    \mathbf{Y}_{i,j,c_{o}} = \sum_{c_{i}=0}^{C_{i}-1}\sum_{k=0}^{H-1} \mathbf{W}_{k,c_{i}}^{H}\cdot (\mathbf{X}_{(k+i){\rm mod}H,j,c_{i}}^{p})
\end{align}
considering $\forall i\in[0,H\!-\!1]$, $\forall j\in[0,W\!-\!1]$, $\forall c_{i} \in[0,C_{i}\!-\!1]$ and $\forall c_{o}\in[0,C_{o}\!-\!1]$. 

\definecolor{CadetBlue2}{RGB}{142, 229, 238}
\begin{algorithm}[t]
\BlankLine
\textcolor{CadetBlue2}{
       \# B: batch size, C: channel number\\
       \# H: height, W: weight\\
       \# fft1d: FFT conduct in one dimension indicated by 'dim'. \\
       \# ifft1d: inverse FFT, normalized with 1/N by in default} \\
        \textbf{def} ParC\_H(x, weight, bias): \\
            \mbox{\qquad}B, C, H, W = x.shape \\
	    \mbox{\qquad}\textcolor{CadetBlue2}{\# periodic extension} \\
	    \mbox{\qquad}x\_cat = torch.cat([x, x[:, :, :-1, :]], dim=-2) \\
	    \mbox{\qquad}\textcolor{CadetBlue}{\# spatial-ParC} \\
	    \mbox{\qquad}x = F.conv2d(x\_cat, weight, bias, padding=0, groups=C)  \\
	    \mbox{\qquad}return x \\
	    
	\textbf{def} Fast\_ParC\_H(x, weight, bias): \\
	    \mbox{\qquad}B, C, H, W = x.shape \\ 
	    \mbox{\qquad}\textcolor{CadetBlue}{\# input FFT} \\
	    \mbox{\qquad}x = fft1d(x, dim=-2) \\
	    \mbox{\qquad}\textcolor{CadetBlue2}{\# weight FFT} \\
	    \mbox{\qquad}weight = fft1d(weight, dim=-2) \\
	    \mbox{\qquad}\textcolor{CadetBlue2}{\# Fourier-ParC} \\
	    \mbox{\qquad}x = x * torch.conj(weight).view(1, C, H, 1) \\
	    \mbox{\qquad}\textcolor{CadetBlue2}{\# output iFFT} \\
	    \mbox{\qquad}x = ifft1d(x, dim=-2).real \\
	    \mbox{\qquad}x = x + bias.view(1, C, 1, 1) \\
	    \mbox{\qquad}return x
    \caption{PyTorch style Pseudo code of ParC-H and Fast-ParC-H. Codes are based on depth-separable versions. (H means horizontal. It could be easily extended to ParC-V by conducting these operations in a vertical dimension.)}
    \label{algorithm_fast_parc} 
\end{algorithm}

\subsection{Fast-ParC: Speed up ParC with its FFT Equivalent Form} \label{subsection:fast_parc}

\begin{table}[t]
    \centering
    \setlength\tabcolsep{14pt}
    \renewcommand\arraystretch{1.5}
    \begin{tabular}{ll}
        \toprule
        \textbf{Operation} & \textbf{Theoretical complexity} \\
        \midrule
        Self-Attention & $\mathcal{O}(CH^{2}W^{2}+C^{2}HW)$\\
        Conv2d & $\mathcal{O}(CHWK_{x}K_{y})$ \\
        ParC   & $\mathcal{O}(CHW(H+W))$ \\
        Fast-ParC & $\mathcal{O}(CHW(\lceil\log_{2}{H}\rceil+\lceil\log_{2}{W}\rceil)$ \\
        \bottomrule
    \end{tabular}
  \caption{The theoretical complexity of multiplication for different basic operations. The complexity of Fast-ParC is computed with base-2-FFT, considering the multiplication of each complex number as equivalent to four real number multiplications. The Detailed complexity computation steps are included in the Appendix.}
  \label{tab:fft_theoratical}
\end{table}

As shown in Table~\ref{tab:fft_theoratical}, the ParC reduces the complexity of global operation from quadratic to 1.5 order, which, however, may still not be enough for inputs of large resolution. \textbf{Could we take a further step towards the linear spatial complexity without sacrificing any performance}? 

One elegant way to explore this concept is to utilize the well-known convolution theorem. In this subsection, we propose an accelerated version of ParC named Fast-ParC to further reduce the computational complexity under large feature resolutions. It is worth noting that Fast-ParC is a complementary implementation form of ParC rather than another new operation. 

\begin{figure*}[t]
    \centering
    \includegraphics[scale=0.58]{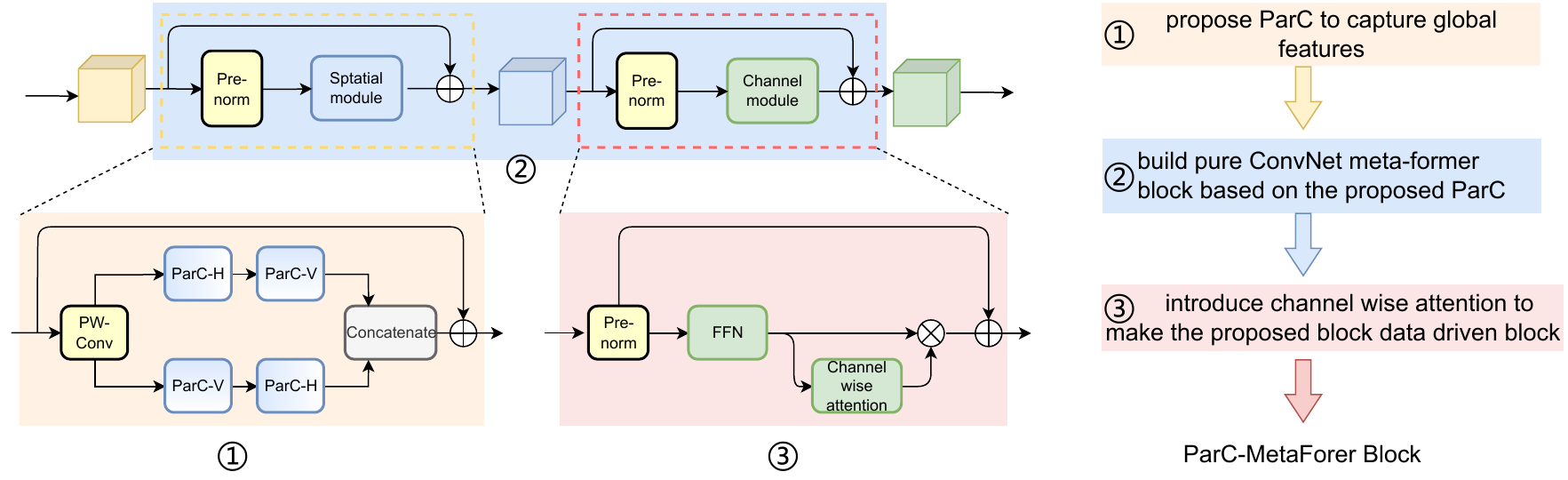}
    \caption{ParC-MetaFormer block used to replace transformer blocks in ViTs or hybrid structures (e.g.     MobileViT~\citep{MobileViT}). ParC-MetaFormer block adopts a MetaFormer~\citep{metaformer} like block structure, and it keeps the three merits of transformer block: 1) global receptive field; 2) positional-aware; 3) data-driven. 
    }
    \label{fig:parc_metaformer_block}
\end{figure*}

We derive Fast-ParC with the help of FFT(Fast Fourier Transform). It is well known that FFT could facilitate convolution operation. According to convolution theorem~\citep{SS_book}, \textit{circular convolution} in the spatial domain is equivalent to the \textit{Hadamard product} in the Fourier domain, which is precisely one of the distinctive differences between ParC and ordinary convolution. 
The convolution theorem, the interesting equation that allows us to develop a neat and beautiful frequency domain implementation for ParC, could be written as:\\
\begin{theorem}{Theorem 1:}{Convolution Theorem}
    \begin{equation} \label{thm:convolution_theorem}
        \begin{aligned}
            & \text{Spatial Form: } &y(n) &= \textstyle\sum_{i=0}^{N-1}w(i)x((n+i))_{N}\\
            & \qquad\Updownarrow & & \\
            & \text{Fourier Form: } &Y(k) &= W^{\ast}(n) X(n)
        \end{aligned}
    \end{equation}
\end{theorem}, in which we define $x(n)$, $w(n)$, $y(n)$ as input, weight, and output sequence in the time domain, $X(k)$, $W(k)$ and $Y(k)$ as the sequence in Fourier domain.

Based on this, we propose a Fourier domain version of the ParC operation named Fast-ParC. Algorithm~\ref{algorithm_fast_parc} shows the pseudo-code to implement Fast-ParC in a PyTorch-style framework. We also proved in theory that Fast-ParC is strictly equivalent to the spatial form of ParC and provided the mathematical proof in the Appendix. This provides ParC with enormous flexibility. Thanks to this strict equivalence, we can choose the more appropriate implementation between ParC/Fast-ParC depending on the current feature resolution and running platform to achieve the largest possible efficiency. 

The advantages of Fast-ParC are clear. Corresponding with Table~\ref{tab:fft_theoratical}, assuming an feature resolution of $N=H=W$, the multiplication complexity of MHA(Multi-Head Attention) is $\mathcal{O}(N^4)$, and ParC in the spatial domain requires $\mathcal{O}(N^{2})$, while Fast-ParC costs $\mathcal{O}(NlogN)$. It's clear that when the feature resolution $N$ is large, the complexity of MHA and ParC will surpass the Fast-ParC by a large margin. 
For example, for COCO~\citep{COCO}, the commonly used resolution for testing is 1280$\times$800, and for ADE20k~\citep{ADE20K} is 2048$\times$512. When N is large, Fast-ParC can save the model's FLOPs and achieve a better acceleration. Another promising direction is that Fast-ParC allows us to use ParC for the shallower stages with an acceptable budget in computation, which is necessary for applying ParC in novel architectures~\citep{NextViT}. 

\subsection{Applying ParC/Fast-ParC on ViTs and ConvNets}
To validate the effectiveness of ParC/Fast-ParC as a plug-and-play meta-operation, we built a series of ParC-based Models based on the operations proposed in Section~\ref{subsection:parc_operation}. Here, baseline models include both ViTs and ConvNets. Specifically, for ViTs, MobileViT~\citep{MobileViT} is selected as the baseline, as it achieved the best parameter/accuracy trade-off among the lightweight hybrid structures proposed recently. For ConvNets, ResNet50~\citep{ResNet}, MobileNetv2~\citep{MobileNetV2} and ConvNext~\citep{ConvNext} are adopted as ConvNet baselines. ResNet50 is the most widely used model in practical applications. MobileNetV2 is the most popular model for mobile devices. ConvNext is the first ConvNet that remains pure ConvNet architecture while integrating some characteristics of ViTs. All of the four models we adopted here are representative. 

\subsubsection{ParC-ViTs}

\begin{figure*}[t]
    \centering
    \includegraphics[scale=0.5]{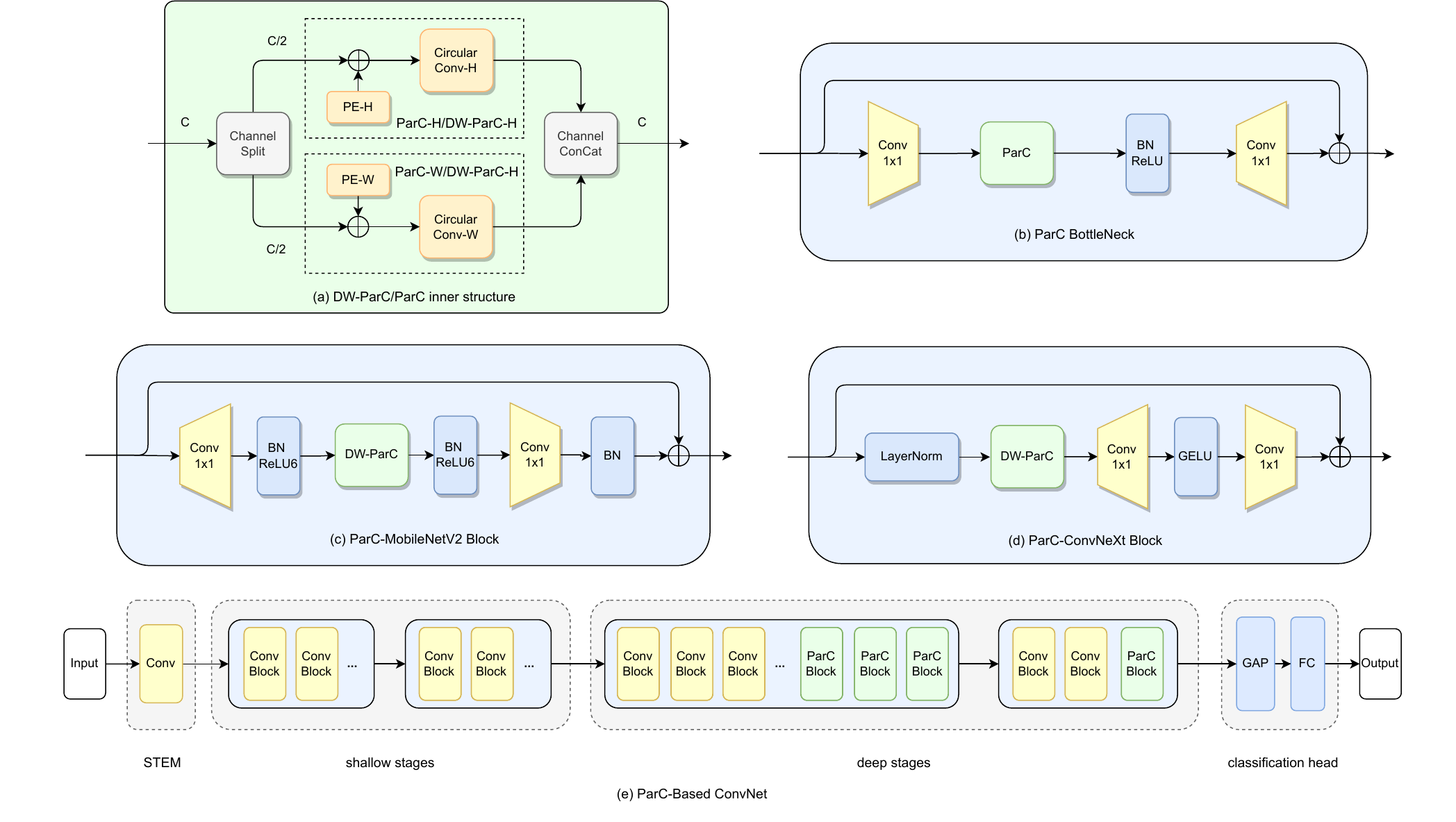}
    \caption{The hierarchical structure of ParC-based ConvNets. Fig.~(a) shows the inner structure of the ParC or Dw-ParC(i.e., depth-wise version ParC) operation. Fig.~(b)(c)(d) illustrates three different ParC-ConvNet blocks. Fig (e) shows the ParC-ConvNet structure, which indicates we replace the last several blocks in the deep stages and keep the structure in the shallow stages.}
    \label{fig:parc_convnet_block_structure}
\end{figure*}

\textbf{ParC-MetaFormer Block.}
As shown in Fig~\ref{fig:parc_metaformer_block} and Fig~\ref{fig:parc_convnet_block_structure}, ConvNets and ViTs have significant differences in the outer layer structure. ViTs generally adopt meta-former blocks as basic architecture. To apply the ParC operation on ViTs, we design the ParC-MetaFormer block to replace transformer blocks in ViTs.

\textbf{Adopting MetaFormer-like structure.}
A MetaFormer~\citep{metaformer} block is the block structure that ViTs use most frequently, and it generally consists of a sequence of two components: a token mixer and a channel mixer. Both two components use residual structure. We adopt ParC as a token mixer to build a ParC-MetaFormer block. We do this because ParC can extract global features and interact information among pixels from global space, which meets the requirement of the token mixer module. And unlike self-attention, whose complexity is quadratic, ParC is much more efficient in computation. Replacing this part with ParC can reduce the computational cost significantly. In the ParC-MetaFormer Block, we adopt a serial structure of ParC-H and ParC-V. Considering symmetry, half of the channels go through ParC-H first, and others go through ParC-V first (as shown in Fig~\ref{fig:parc_metaformer_block}). 

\textbf{Adding channel-wise attention in channel mixer part.}
Though ParC keeps the global receptive field and positional embedding, another benefit ViTs have against ConvNets is data-driven. In ViTs, self-attention module can adapt weights according to input. This makes ViTs data-driven models, which can focus on essential features and suppress unnecessary ones, bringing better performance. Previous literature~\citep{SENet}~\citep{CBAM}~\citep{micronet} already explained the importance of keeping the model data-driven. By replacing the self-attention with the proposed global circular convolution, we get a pure ConvNet that can extract global features. However, the replaced model is no longer data-driven. To compensate, we insert a channel-wise attention module into the channel mixer part, as shown in Fig.~\ref{fig:parc_metaformer_block}. Following SENet~\citep{SENet}, we first aggregate spatial information of input features $x\in\mathbb{R}^{c\times h\times w}$ via global average pooling and get aggregated feature $x_{a}\in\mathbb{R}^{c\times1\times1}$ Then we feed $x_{a}$ into a multi-layer perception to generate channel wise weight $a\in\mathbb{R}^{c\times 1\times 1}$ The $a$ is multiplied with $x$ channel wise to generate the final output.

\textbf{MobileViT-ParC Network.}
Currently, existing hybrid structures can be divided into three main kinds of structures, including serial structure~\citep{LeViT}~\citep{VIT-C}, parallel structure~\citep{MobileFormer}, and bifurcate structure~\citep{MobileViT}~\citep{CoAtNet}. Among all three structures, the third one achieves the best performance for now. MobileViT~\citep{MobileViT} also adopts the bifurcate structure. Inspired by this, based on MobileViT, we also build our model with a bifurcate structure. MobileViT consists of two major types of modules. Shallow stages consist of MobileNetV2 blocks, which have a local receptive field. Deep stages comprise ViT blocks, which enjoy a global receptive field. We keep all MobileNetV2 blocks and replace all ViT blocks with corresponding ParC blocks. This replacement converts the model from a hybrid structure to pure ConvNet while reserving its global feature-extracting ability.

\begin{table*}[ht]
    \centering
    \setlength\tabcolsep{10pt}
    \begin{tabular}{lllll}
    \toprule
    Frameworks & Models & Source & \makecell{Parmas (M)} & \makecell{Top1 Accu (\%)} \\
    \midrule
    \multirow{4}{*}{Pre-ConvNet} 
        & SuffleNetV2(2.0x) & ECCV 2018 & 5.5  & 74.5 \\
        & MobileNetV3(1.0x) & ICCV 2019 & 5.4  & 75.2 \\
        & EfficientNet-B0   & ICML 2019 & 5.3  & 76.3 \\
        & ResNet-101-SE     & CVPR 2018 & 49.3 & 77.6 \\
    \midrule
    \multirow{2}{*}{ViTs} 
        & DeiT-2G & ICML 2021 & 9.5  & 77.6 \\
        & Swin-1G & ICCV 2021 & 7.3  & 77.3 \\
    \midrule
    \multirow{6}{*}{\makecell{Hybrid Structures\\(ConvNet+ViT)}}
        & CoaT-Lite-T        & ICCV 2021  & 5.7  & 76.6 \\
        & LeViT-128S         & ICCV 2021  & 7.8  & 76.6 \\
        & Mobile-Former      & CVPR 2022  & 9.4  & 76.7 \\
        &  $\text{MobileViT-S}^{\dagger}$       & ICLR 2022  & 5.6  & 78.4 \\
        &  $\text{EfficientFormer-L1}^{\dagger}$ & NeurIPS 2022 & 12.3 & 76.0 \\
    \midrule
    \multirow{2}{*}{\makecell{Post-ConvNet}}
        &  $\text{ConvNext-XT}^{}$        & CVPR 2022  & 7.4  & 77.5 \\
        & $\text{MobileViT-ParC-S}^{\dagger}$        & ECCV 2022  & \textbf{5.0} & \textbf{78.6} \\
    \bottomrule
    \end{tabular}
    \caption{Classification results on ImageNet-1K. $\dagger$
    means running in MobileViT\cite{MobileViT} training framework.
    Pre-ConvNet indicates classical ConvNets before ViTs appeared, Post-ConvNet integrates the merits of both ViTs and ConvNets.}
    \label{transformer_imagnet_result}
\end{table*}

\subsubsection{ParC-ConvNets}
For ParC-ConvNets, we focus on providing ConvNets with a global receptive field. Replacing vanilla convolution with ParC operation (as shown in Fig.~\ref{fig:parc_convnet_block_structure}~(a)), we build different ParC-based blocks (as shown in Fig.~\ref{fig:parc_convnet_block_structure}~(b) (c)(d)). Previous hybrid structure works~\citep{LeViT}~\citep{VIT-C}~\citep{MobileViT} draw a similar conclusion: models with local-extracting blocks in early stages and global-extracting blocks in deep stages achieve the best performance. Since ParC owns a global receptive field, we insert a ParC-based block into ConvNets following this rule (as shown in Fig.~\ref{fig:parc_convnet_block_structure}~(e)).

\textbf{ParC BottleNeck and ResNet50-ParC Network.} ResNet~\citep{ResNet} is one of the most classical ConvNet. Simply replacing the 3$\times$3 convolution of the original ResNet50 BottleNeck with a ParC operation, we can obtain the ParC BottleNeck (see Fig~\ref{fig:parc_convnet_block_structure}~(b)). Since the characteristics of ParC-H and ParC-V might be significantly different, no channel interaction is introduced between them. This is similar to adopting group convolution~\citep{ResNeXt} with group=2. 
The central part of ResNet can be divided into four stages, each consisting of a couple of repeated BottleNeck blocks. Specifically, ResNet50 has [3, 4, 6, 3] blocks in four stages. By replacing the last $1/2$ of the 3rd stage of ResNet50 and the last $1/3$ of the 4th stage with ParC BottleNeck, we obtain the ResNet50-ParC. 

\textbf{ParC-MobileNetV2 Block and MobileNetV2-ParC Network.}
MobileNetV2~\citep{MobileNetV2} is a typical representative of the lightweight model. By replacing the 3$\times$3 depthwise convolution in Inverted Bottleneck with depthwise ParC, we get ParC-MobileNetV2 block (see Fig~\ref{fig:parc_convnet_block_structure}~(c)).
MobileNetV2 is much slimmer and deeper than Resnet50, with blocks number of [1, 2, 3, 4, 3, 3, 1] respectively for seven stages. MobilenetV2-ParC could be obtained by replacing the last $1/2$ blocks of stage 4 and the last $1/3$ blocks of stages 5 and 6 with the ParC-MobileNetV2 blocks. 

\textbf{ParC-ConvNeXt Block and ConvNeXt-ParC Network.}
ConvNeXt~\citep{ConvNext} takes a sequence of modifications on the original ResNet50 structure to learn from transformers. During this, 3$\times$3 convolution is replaced with 7$\times$7 depthwise convolution. This enlarges the local receptive field; however, it still cannot grad global information. We further replace 7$\times$7 depthwise convolution in ConvNeXt Block by depthwise ParC. By this, we obtain the ParC-ConvNeXt block ( Fig~\ref{fig:parc_convnet_block_structure}~(d)). 
Replacing the last $1/3$ blocks of the last two stages of ConvNeXt with ParC-ConvNeXt blocks, an example of ConvNeXt-ParC is obtained. We reduce the basic channel number in ConvNeXt-T to 48 (i.e. [48, 96, 192, 384] for each stage) to get a lightweight ConvNeXt-XT, which is more welcome in deploying on edge computing devices, and it also has a shorter experimental period.   

Note that, in ParC-MetaFormer, a sequence of ParC-H and ParC-W are adopted to keep the receptive field consistent with self-attention, since this design is used to replace self-attention. In ParC-ConvNets, we adopt a parallel of ParC-H and ParC-V (every single layer) as shown in Fig.~\ref{fig:parc_convnet_block_structure}. According to experimental results, this setting could already provide enough performance gain compared to the vanilla ConvNets. In fact, as we do not use only one ParC-ConvNet block, the ParC-ConvNets still have a global receptive field.

\section{Experiment}
To test the capability of ParC as a plug-and-play operation. We applied it to a wide range of models, including Transformers and ConvNets. We take different views when plugging ParC into these two kinds of models. For transformer models, the main purpose of applying ParC is to make the model more hardware-friendly while keeping its ability of global extracting. Thus, we use ParC to replace self-attention. For ConvNets, we use ParC mainly to enable networks with global extracting ability while keeping it parameter/FLOPs friendly. Thus, we use ParC to replace vanilla convolution. Notice we not only verify the performance of our models on classification tasks but also on downstream tasks (e.g., object detection, semantic segmentation). This comprehensive approach is crucial, as previous research \citep{PVT, harmful_patch} has indicated that superior classification accuracy does not necessarily guarantee good performance on other downstream tasks.

\subsection{Experiment on Vision Transformer Models} \label{subsection:experiment_vit}
In the transformer experiments, we focus on inserting the ParC operation into the most recently proposed MobileViT framework. We show the overall advantages of the proposed MobileViT-ParC on three typical vision tasks, including image classification, object detection, and semantic segmentation.

\subsubsection{Image Classification on ImageNet-1K}
We conduct image classification experiments on ImageNet-1k, the most widely used benchmark dataset for this task. We train the proposed MobileViT-ParC models on the training set of ImageNet-1K and report top-1 accuracy on the validation set.

\textbf{Training setting.}
As we adopt a MobileViT-like structure as our outer framework, we train our model using a very similar training strategy as well. To be specific, we train each model for 300 epochs on 8 V100 or A100 GPUs with AdamW optimizer~\citep{adamw}, where the maximum learning rate, minimum learning rate, weight decay, and batch size are set to 0.004, 0.0004, 0.025, and 1024 respectively. Optimizer momentum $\beta_{1}$ and $\beta_{2}$ of the AdamW optimizer are set to 0.9 and 0.999, respectively. We use the first 3000 iterations as the warm-up stage. We adjust the learning rate following the cosine schedule. For data augmentation, we use random cropping, horizontal flipping, and multi-scale samplers. We use label smoothing~\citep{label_smoothing} to regularize the networks and set the smoothing factor to 0.1. We use Exponential Moving Average (EMA)~\citep{ema}. More details of the training settings and links to the source code will be provided in supplementary materials.

\textbf{Results comparison.}
The experiment results of image classification and comparison with other models are listed in Table~\ref{transformer_imagnet_result}. Table~\ref{transformer_imagnet_result} shows that MobileViT-ParC-S and MobileViT-S beat other models by a clear margin. The proposed MobileViT-ParC-S achieves the highest classification accuracy and has fewer parameters than most models. Compared with the second-best model, MobileViT-S, our MobileViT-ParC-S decreases the number of parameters by 11\% and increases the top-1 accuracy by 0.2 percent. Lightweight models. Firstly, comparing the results of lightweight ConvNets with that of ViTs, lightweight ConvNets show much better performance. Secondly, comparing the popular ConvNets before ViT appears (pre-ConvNets), ViTs, and hybrid structures, hybrid structures achieve the best performance. Therefore, improving ConvNets by learning from the merits of ViT is feasible. Finally, the proposed MobileViT-ParC achieves the best performance among all comparison models. So, indeed, by learning from ViT design, the performance of pure lightweight ConvNets can be improved significantly.

\begin{table*}
    \begin{minipage}[b]{0.499\linewidth}
        \centering
        \setlength\tabcolsep{14pt}
        \begin{tabular}{lll}
            \toprule
            Models & \makecell{Params (M)} & mAP \\
            \midrule
            MobileNetV2      & 4.3  & 22.1 \\
            MixNet           & 4.5  & 22.3 \\
            MNASNet          & 4.9  & 23.0 \\
            MobileViT        & 2.7  & 24.8 \\
            VGG              & 35.6 & 25.1 \\
            ResNet50         & 22.9 & 25.2 \\
            \midrule
            MobileViT-S      & 5.7 & 27.7 \\
            MobileViT-ParC-S & \textbf{5.2} & \textbf{28.8} \\
            \bottomrule
        \end{tabular}
        \par\vspace{0pt}
    \end{minipage}%
    \begin{minipage}[b]{0.499\linewidth}
        \centering
        \setlength\tabcolsep{14pt}
        \begin{tabular}{lll}
            \toprule
            Models & \makecell{Params (M)} & mIoU \\
            \midrule
            MobileNetV1       & 11.2  & 75.3 \\
            MobileNetV2       & 4.5   & 75.7 \\
            MobileViT-XXS     & 1.9   & 73.6 \\
            MobileViT-XS      & 2.9   & 77.1 \\
            ResNet101         & 22.9  & 80.5 \\
            \midrule
            MobileViT-S      & 6.4 & 79.1 \\
            MobileViT-ParC-S & \textbf{5.8} & \textbf{79.7} \\
            \bottomrule
        \end{tabular}
        \par\vspace{0pt}
    \end{minipage}
    \caption{Object detection results on MS-COCO (Left) and Semantic segmentation experiments on PASCAL VOC (Right) about ParC-based transformer models. We compare the mAP/mIoU with other models.}
    \label{transformer_downsteam}
\end{table*}

\subsubsection{Object detection on MS-COCO}
We use MS-COCO~\citep{MS_COCO} datasets and its evaluation protocol for object detection experiments. Following~\citep{MobileNetV2}~\citep{MobileViT}, we take single-shot object detection (SSD)~\citep{SSD} as the detection framework and use separable convolution to replace the standard convolutions in the detection head.

\textbf{Experiment setting.}
With models pre-trained on ImageNet-1K as the backbone, we fine-tune detection models on the training set of MS-COCO with AdamW optimizer for 200 epochs. Batch size and weight decay are set to 128 and 0.01. We use the first 500 iterations as a warm-up stage, where the learning rate is increased from 0.000001 to 0.0009. Both label smoothing and EMA are used during training.

\textbf{Results comparison.}
Table~\ref{transformer_downsteam} lists the corresponding results. Similar to results in image classification, MobileViT-S and MobileViT-ParC-S achieve the second best and the best in terms of mAP. Compared with the baseline model, MobileViT-ParC-S shows advantages in both model size and detection accuracy.

\subsubsection{Semantic segmentation on VOC/COCO}
\textbf{Experiment setting.}
DeepLabV3 is adopted as the semantic segmentation framework. We fine-tune models on the training set of PASCAL VOC~\citep{VOC} and COCO~\citep{COCO} dataset, then evaluate trained models on the validation set of PASCAL VOC using mean intersection over union (mIoU) and report the final results for comparison. We fine-tune each model for 50 epochs with AdamW.

\begin{table}[t]
    \centering
    \setlength\tabcolsep{1pt}
    \begin{tabular}{lccc}
        \toprule
        Models & Parms (M) & FLOPs (G) & Top1 Accu (\%) \\
        \midrule
        $\text{ResNet50}^{}$ & 25.6  & 4.1 & 79.1 \\ 
        $\text{ResNet50-ParC}^{}$ & \textbf{23.7} & \textbf{4.0} &  \textbf{79.6}\\
        \midrule
        $\text{MobileNetV2}^{}$ & 3.51 & 0.6 & 70.2 \\ 
        $\text{MobileNetV2-ParC}^{}$  & \textit{3.54} & 0.6 & \textbf{71.1} \\
        \midrule       
        $\text{MobileNetV2}\dagger^{}$ & 3.51 & 0.6 & 72.7 \\ 
        $\text{MobileNetV2-ParC}\dagger^{}$  & \textit{3.54} & 0.6 & \textbf{73.3} \\
        \midrule
        $\text{ConvNeXt-XT}^{}$ & 7.44 & 1.1 & 77.5\\ 
        $\text{ConvNeXt-ParC-XT}^{}$ & \textbf{7.41} & 1.1 & \textbf{78.3} \\
        \bottomrule
    \end{tabular}
    \caption{Classification results on ImageNet-1K about ParC based ConvNets. Most of the models are running in ConvNext's~\cite{ConvNext} training recipe, while with $\dagger$, which are trained with the recipe recommended by torchvision\cite{pytorch} for MobileNetV2. }
    \label{convnet_imagenet_result}
\end{table}

\begin{table*}[t]
  \centering
    \setlength\tabcolsep{0.5pt}
    \begin{tabular}{llllllllllllll}
    \toprule
    \multicolumn{2}{c}{Backbone} & \multicolumn{6}{c}{Instance Detection} & \multicolumn{6}{c}{Instance Segmentation} \\
    \cmidrule(lr){1-2}\cmidrule(lr){3-8}\cmidrule(lr){9-14}
    Model & \makecell{Params (M)} & $\text{AP}^{\text{box}}$ & $\text{AP}_{\text{50}}^{\text{box}}$ & $\text{AP}_{\text{75}}^{\text{box}}$ & $\text{AP}_{\text{S}}^{\text{box}}$ & $\text{AP}_{\text{M}}^{\text{box}}$ & $\text{AP}_{\text{L}}^{\text{box}}$ & $\text{AP}^{\text{mask}}$ & $\text{AP}_{\text{50}}^{\text{mask}}$ & $\text{AP}_{\text{75}}^{\text{mask}}$ & $\text{AP}_{\text{S}}^{\text{mask}}$ & $\text{AP}_{\text{M}}^{\text{mask}}$ & $\text{AP}_{\text{L}}^{\text{mask}}$ \\
    \midrule
    ResNet50 & 25.6 & 47.5 & 65.6 & 51.6 & 29.9 & 51.1 & 61.5 & 41.1 & 63.1 & 44.6 & 23.6 & 44.7 & 55.3  \\
    ResNet50-ParC & \textbf{23.7} & \textbf{48.1} & \textbf{66.4} & \textbf{52.3} & \textbf{30.3} & \textbf{51.2} & \textbf{62.3} & \textbf{41.8} & \textbf{64.0} & \textbf{45.1} & \textbf{24.6} & \textbf{45.2} & \textbf{56.0} \\
    \midrule
    MobileNetV2 & 3.51 & 43.7 & 61.9 & 47.6 & 27.0 & 46.1 & 58.3 & 37.9 & 59.1 & 40.8 & 21.4 & 40.5 & 52.5 \\
    $\text{MobileNetV2-ParC}^{\dagger}$ & \textit{3.54} & \textbf{44.3} & \textbf{62.7} & \textbf{47.8} & \textbf{28.0} & \textbf{47.0} & \textbf{58.6} & \textbf{39.0} & \textbf{60.3} & \textbf{42.1} & \textbf{23.0} & \textbf{41.8} & \textbf{53.1}  \\
    \midrule
    ConvNeXt-XT & 7.44 & 47.2  & 65.6 & 51.4 & 30.1 & 50.2 & 62.3 & 41.0 & 62.9 & 44.2 & 23.8 & 44.2 & 55.9 \\
    ConvNext-ParC-XT& \textbf{7.41} & \textbf{47.7} & \textbf{66.2} & \textbf{52.0} & \textbf{30.4} & \textbf{50.9} & \textbf{62.7} & \textbf{41.5} & \textbf{63.6} & \textbf{44.6} & \textbf{24.4} & \textbf{44.7} & \textbf{56.5}  \\
    \bottomrule
    \end{tabular}
  \caption{Instance detection and segmentation results on COCO dataset of convolution models and ParC-based convolution models. The code is based on MMdetection. $\dagger$ means a 3$\times$ expansion of learnable parameters to ParC weight is applied when transferred from ImageNet (i.e. size 224$\times$224) to COCO (i.e. size 800$\times$1333).}
  \vspace{-0.6cm}
  \label{convnet_COCO}
  
\end{table*}

\textbf{Results comparison.}
Results are summarized in Table~\ref{transformer_downsteam}. We can see that MobileViT-S and MobileViT-ParC-S have the best trade-off between model scale and mIoU. Compared with ResNet-101, MobileViT-S, and MobileViT-ParC-S achieve competitive mIoU while having much fewer parameters.

\subsection{Experiment on Convolutional Models} \label{subsection:experiment_convnet}
In the convolutional networks experiments, we insert the ParC module into ResNet50, MobileNetV2, and ConvNeXt and then examine their performance improvement compared with their baselines. These parts of experiments also include the three typical vision tasks: classification, detection, and segmentation. 
Since ViT models~\citep{ViT}~\citep{Swin} achieved great performance, and the training recipes they shared (e.g., larger batch size, longer epochs) have been proven to be beneficial for ConvNets as well~\citep{ConvNext}, we train and evaluate all of the aforementioned convolutional models with ConvNeXt~\citep{ConvNext} training recipes with their origin network architectures to get better performance. The only modification in network architecture is the application of ParC.

\subsubsection{Image Classification on ImageNet-1K} \label{subsubsection:experiment_convnet_imagenet}
\textbf{Training setting.} 
Most of our training settings for convolutional models are from ConvNeXt's~\citep{ConvNext} guide. To be specific, we use 8 2080Ti or 3090 to train each model for 300 epochs with the AdamW optimizer. The learning rate increases linearly in the beginning 20 epochs as the warm-up and then decays with the cosine schedule. The batch size, base learning rate, weight decay, and momentum $\beta_{1}$ and $\beta_{2}$ are set as 4096, 0.004, 0.05, 0.9, and 0.999, respectively. Data augmentations used include Mixup, Cutmix, RandAugment, and Random Erasing. Regularization methods used include Stochastic Depth~\citep{stochastic_depth} and Label Smoothing~\citep{label_smoothing}. Notice in this section that the EMA(Exponential Moving Average) skill is NOT used since, in most experiments, we observed that the original models achieve higher accuracy than their EMA counterparts. 

For MobileNetV2, since we observed an unexpected performance decline after applying the training recipe of ConvNeXt. We add another group of compassion with the commonly used recipe suggested by Torchivison ~\citep{pytorch}. The total training epochs are increased to 600, with a 5-epoch linear warm-up. The cosine annealing is used as the scheduler. The batch size, base learning rate, and weight decay are set as 128, 0.5, and 0.00001. Data augmentations used include Mixup, Cutmix, and Random Erasing.

\textbf{Results comparison.}
We show the results of different types of ConvNets on ImageNet-1K in Table~\ref{convnet_imagenet_result}. It's clear that all three ParC-enhanced models beat their prototypes. Especially for ResNet50, using ParC design improves classification accuracy by 0.5\% while reducing 2M parameters and saving computation costs slightly. Comparing the lightweight model MobileNetV2 and MobileNetV2-ParC, our operation shows an extremely obvious advantage with an improvement of 0.9\%, with a slight cost on latency. ConvNeXt-ParC-XT exceeds the original ConvNeXt-XT by 0.8\%, too. 

Compared with the result reported in its original paper~\citep{ResNet}, the ConvNeXt training recipe helped improve ResNet50's performance. However, the torchvision training recipe seems to benefit MobileNetV2 more, with an evaluation accuracy of 72.7\% (even surpassing the official result of 72.2\%), which is better than the ConvNeXt recipe of only 70.2\%. Under this recipe, the application of ParC still boosts MobileNetV2's performance by 0.6\%. 

Hence, generally speaking, ParC-based convolutional models achieve higher accuracy with almost the same (or even fewer) FLOPs and parameters as the original models. 

\subsubsection{Instance detection and segmentation on COCO}
\textbf{Training setting.} 
We use ConvNets pre-trained on ImageNet (in Section~\ref{subsubsection:experiment_convnet_imagenet}) as backbones and then fine-tune with Cascade Mask-RCNN~\citep{cascade_mask_rcnn} as detection framework on COCO dataset~\citep{COCO}. We set epoch as 36 and learning rate as 2e-4 (decay by 10 at epochs 27 and 33). We set momentum $\beta_{1}$ and $\beta_{2}$ as 0.9 and 0.999, weight decay as 0.05, and stochastic depth as 0.4. The code is built based on the official implementation of MMDection~\citep{mmdetection} toolboxes. FP16 training is NOT used for precision concerns. While training ResNet50 and ResNet50-ParC, we froze the first stage of the models. To train ConvNeXt-XT and ConvNeXt-ParC-XT, we set the basic layer-wise learning rate decay to 0.7. For training networks with BN(Batch Normalization) layers, we set BN layers to evaluation mode throughout the whole process. We use multi-scale training on the training set and report the box AP and mask AP on the validation set. More detailed training and testing configurations can be seen in our repository.

For MobileNetV2-ParC, since the resolution amplification in our training recipes is about 3 times from ImageNet (224$\times$224) to COCO (800$\times$1333). Specifically, taking ParC kernels in stage 3 as an example, we extend 14$\times$1-sized ParC-H kernels and 1$\times$14-sized ParC-W kernels to 21$\times$1 and 1$\times$42, respectively. We extend kernels pre-trained on ImageNet-1K to initialization kernels for detection and segmentation models. 

\textbf{Results comparison.}
The result of instance detection and segmentation is listed in Table~\ref{convnet_COCO}, which is almost consistent with the classification result on ImageNet. The ParC-based models outperform the vanilla models by a clear margin in both box AP and mask AP. Specifically, ResNet50-ParC improves by 0.6 in box AP and 0.7 in mask AP, MobileNetV2-ParC improves by 0.6 in box AP and 1.1 in mask AP, ConvNeXt-ParC-XT improves by 0.5 in both box AP and mask AP. 

\begin{table*}[t]
    \centering
    \setlength\tabcolsep{11.5pt}
    \begin{tabular}{llllllll}
        \toprule
        \multicolumn{2}{c}{Backbone} & \multicolumn{3}{c}{Evaluation w/o TTA} & \multicolumn{3}{c}{Evaluation w/ TTA} \\
        \cmidrule(lr){1-2}\cmidrule(lr){3-5}\cmidrule(lr){6-8}
        Model & \makecell{Params (M)} & mIoU & mAcc & aAcc & $\text{mIoU}_{\text{aug}}$ & $\text{mAcc}_{\text{aug}}$ & $\text{aAcc}_{\text{aug}}$ \\
        \midrule
        ResNet50 & 25.6  & 42.27 & 52.91 & 79.88 & 43.75& 53.50 & 80.71 \\
        ResNet50-ParC & \textbf{23.7} & \textbf{44.32} & \textbf{54.66} & \textbf{80.80} & \textbf{44.69} & \textbf{54.38} & \textbf{81.33} \\
        \midrule
        MobileNetV2 & 3.51  & 38.66 & 48.75  & 77.98 & 39.26 & 48.27 & 78.87 \\
        $\text{MobileNetV2-ParC}^{\dagger}$ & \textit{3.54}  & \textbf{39.25} & \textbf{49.64} & \textbf{78.35} & \textbf{39.62} & \textbf{48.57} & \textbf{79.03} \\
        \midrule
        ConvNeXt-XT & 7.44  & 42.17 & 54.18 & 79.72 & 43.07 & 54.27 & 80.44 \\
        ConvNeXt-ParC-XT & 7.41 & \textbf{42.32} & \textbf{54.48} & \textbf{80.30} & \textbf{43.09} & \textbf{54.41} & \textbf{80.76} \\
        \bottomrule
    \end{tabular}
    \caption{Sementic segmentation result on ADE20k dataset of convolution models and ParC-based convolution models. Test Time Argumentation(TTA) schemes used by us include multi-scaling of [0.5, 1.75] and flipping. $\dagger$ means a 4$\times$ expansion of learnable parameters to ParC weight is applied when transferred from ImageNet (i.e. size 224$\times$224) to ADE20k (i.e. size 512$\times$2048).}
    \vspace{-0.6cm}
    \label{convnet_segm_ADE20K}
\end{table*}

\subsubsection{Semantic segmentation on ADE20K}
\textbf{Training setting.} 
We use a convolutional model pre-trained on ImageNet as the backbone and fine-tune with UperNet~\citep{UPernet} as the framework on the ADE20K~\citep{ADE20K} dataset. We set the max iteration to 16000. We set the basic learning rate as 1, using a linear warm-up and the poly-deacy schedule. We set the momentum, $\beta_{1}$, and $\beta_{2}$ to 0.9, 0.9 and 0.999, respectively. Weight decay and stochastic depth are set to 0.05 and 0.4. The code is built based on the official implementation of MMSegmentation~\citep{mmsegmentation} toolboxes. F16 training is \textbf{NOT} used for precision concerns. For networks with BN, setting all BN layers to evaluation mode during fine-tuning. We froze the first stage of ResNet50 and ResNet50-ParC. For ConvNeXt-XT and ConvNeXt-ParC-XT, we use a basic layer-wise learning rate decay of 0.9.
For MobileNetV2-ParC, we extend the learnable parameters of ParC to its 4$\times$ times and use the interpolation result as initialization. We do this adaptation step because the resolution of ADE20k (i.e., 512$\times$2048) is larger than ImageNet (i.e., 224$\times$224). 
We use multi-scale training and report mIoU, mAcc, and aAcc results on the validation set. Following ConvNeXt~\citep{ConvNext} and Swin~\citep{Swin}, we enable 'slide' mode during evaluation. By doing this, the original picture would be cut into several patches and passed to the network individually. Specifically, we set crop size as 512$\times$512 and stride as 341 (take the average for overlapping area). We test the original single-scale mIoU and also the mIoU with test time argumentation (TTA). The TTA used include: 1) flipping; 2) multi-scaling of range [0.5, 1.75]. Results are shown in Table~\ref{convnet_segm_ADE20K}.

\begin{figure*}[t]
    \centering
    \includegraphics[scale=0.5]{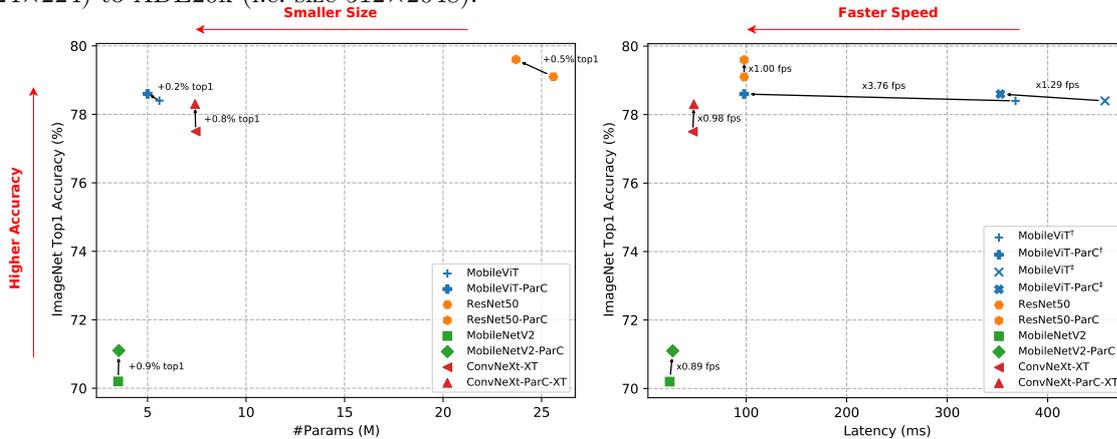}
    \caption{The parameters-accuracy trade-off (left) and latency-accuracy trade-off (right) of ParC-based models and their vanilla counterparts. Arrows in the left figure indicate the accuracy changes, in the right figure indicate the latency changes. $\dag$ and $\dagger$ indicate results measured on different platforms (see Table~\ref{latency_result} for detail.) Figures show that: 1) Replacing self-attention with ParC decreases ViTs' parameters/latency drastically, with a slight improvement in performance; 2) Replacing convolution with ParC improves the performance over a wide range of ConvNets, 
    while barely increasing the models' parameters/latency.
    }
\label{fig:parc_tradeoff}
\end{figure*}

\textbf{Results comparison.}
The ParC models improve a lot to be compared with the original models. Inserting ParC operation improves about 1.15 in ConvNeXt-XT. In traditional networks, it has an even better performance. Especially, ResNet50-ParC outperforms the original ResNet50 by 2.05, and MobileNetV2-ParC outperforms MobileNetV2 by 0.59 in mIoU. TTA improves performance for all ConvNet models tested here. But, it is clear that vanilla ConvNets could get more benefits from it. Comparing rows 3-5 with rows 6-8 in Table~\ref{convnet_segm_ADE20K}, we could conclude that ParC's advantage degrades significantly after TTA. Even so, ParC version models still show better performance than the vanilla version.

Another interesting fact is that this significant performance gain is closely related to the '512$\times$512-sized crop' strategy during evaluation. If we set the test mode to 'whole', which means feeding the whole 512$\times$2048-sized picture to the network, we could also observe a degradation of ParC-based models' advantage. This is in accord with what we see when enabling TTA. Based on these results, we conclude one possible explanation - ParC is not robust enough to resolution change. While using ImageNet-1K for pre-training, we empirically set the meta-kernel size to 14 or 7 and used bi-linear interpolation to gain a global kernel during the practical forward process. Though the interpolated kernel works, it might be a sub-optimum technique, causing instability in resolution changes. A more promising technique is to generate the kernel dynamically. And since zero-padding convolution does not change the resolution of the feature map, it might be a candidate for the kernel generation. But we, unfortunately, failed to accomplish such a version with better performance than Static-ParC, and this might be related to normalization or optimization problems. But after all, Dynamic-ParC is still a promising direction for future work.

\begin{table*}[t]
    \centering
    \setlength\tabcolsep{10pt}
    \begin{tabular}{lccccc}
        \toprule
        Model & \makecell{Parms (M)} & \makecell{FLOPs (G)} & Devices & \makecell{Speed (ms)} & \makecell{Top1 Accu (\%)} \\
        \midrule
        MobileViT-S        & 5.6 & 4.0 & RK3288 & 457 & 78.4 \\
        MobileViT-ParC-S   & 5.0 & 3.5 & RK3288 & \textbf{353} & 78.6 \\
        \midrule
        MobileViT-S        & 5.6 & 4.0 & DP2000 & 368 & 78.4 \\
        MobileViT-ParC-S   & 5.0 & 3.5 & DP2000 & \textbf{98} & 78.6 \\
        \midrule
        $\text{ResNet50}^{\ast}$       & 26 & 4.1 & CPU & 98 & 79.1 \\ 
        $\text{ResNet50-ParC}^{\ast}$  & 24 & 4.0 & CPU & 98 & 79.6 \\
        \midrule
        $\text{MobileNetV2}^{\ast}$      & 3.5 & 0.6 & CPU & 24 & 70.2 \\
        $\text{MobileNetV2-ParC}^{\ast}$ & 3.5 & 0.6 & CPU & 27 & 71.1 \\
        \midrule
        $\text{ConvNeXt-XT}^{\ast}$      & 7.4 & 1.1 & CPU & 47 & 77.5 \\
        $\text{ConvNeXt-ParC-XT}^{\ast}$ & 7.4 & 1.1 & CPU & 48 & 78.3 \\
        \bottomrule
    \end{tabular}
    \caption{Applying ParC-Net designs on different backbones and comparing inference speeds of different models. CPU used here is Xeon E5-2680 v4, DP2000 is the code name of a in-house unpublished low-power neural network processor that highly optimizes convolutions. $\ast$ denotes the models are trained under ConvNeXt\cite{ConvNext} hyperparameters settings, which may not be the optimal. 
    Latency is measured with batch size 1.}
    \label{latency_result}
\end{table*}
\subsection{Inference Speed Test}
In this section, we offer the latency experiments of ParC-based networks measured on multiple platforms. These experiments are conducted to demonstrate two facts:
\begin{enumerate}
    \item ParC could significantly reduce inference latency when applied to transformer models. To demonstrate this, we deploy MobileViT-ParC on a widely used low-power chip Rockchip RK3288 and an in-house low-power neural network processor DP2000. We use ONNX~\citep{ONNX} and MNN~\citep{MNN} to port these models to chips and time each model for 100 iterations to measure the average inference speed. The latency result is then compared with the vanilla MobieViT.
    \item ParC could significantly boost models' performance with a negligible increase in complexity when applied to convolutional models. To demonstrate this, we measure the parameters, FLOPs, and latency of ParC-based ConvNets on the Xeon E5-2680 v4 CPU. The result is then compared with their vanilla counterparts.
\end{enumerate}

\begin{table*}[t]
    \centering
    \setlength\tabcolsep{12pt}
    \begin{tabular}{lllcccc}
    \toprule
    Row & Task & Kernel & CA & PE & \makecell{Params\\(M)} & Top1/mAP/mIoU\\
    \midrule
    1  & Classification & Baseline & - & - & 5.6 & 78.35 \\
    2  & Classification & BK $L/4$ & T & F & 5.0 & 78.45 \\
    3  & Classification & BK $L/2$ & T & F & 5.0 & 78.46 \\
    4  & Classification & ParC     & F & F & 5.3 & 78.50 \\
    5  & Classification & ParC     & T & F & 5.0 & \textbf{78.63} \\
    6  & Classification & ParC     & T & T & 5.0 & \textbf{78.63} \\
    \midrule
    7  & Detection      & Baseline & - & - & 5.7 & 27.70 \\
    8  & Detection      & ParC     & T & F & 5.7 & 27.50 \\
    9  & Detection      & ParC     & T & T & 5.7 & \textbf{28.50} \\
    \midrule
    10  & Segmentation  & Baseline & - & - & 6.4 & 79.10 \\
    11 & Segmentation   & ParC     & T & F & 5.8 & 79.20 \\
    12 & Segmentation   & ParC     & T & T & 5.8 & \textbf{79.70} \\
    \bottomrule
    \end{tabular}
    \par\vspace{0pt}
    \caption{Ablation study conducted under the framework of MobileViT \citep{MobileViT}. BK $L/n$ denotes the big kernel with a length of $1/n$ times $L$ (resolution), CA denotes channel-wise attention and PE denotes position embedding.}
    \label{ablation_study_mobilevit}
\end{table*}%
\begin{table*}[t]
    \centering
    \setlength\tabcolsep{12pt}
    \begin{tabular}{lllcccc}
    \toprule
    Row & Task & Kernel & CC & PE & \makecell{Params\\(M)} & Top1/mAP/mIoU\\ 
    \midrule
    1 & Classification &   Baseline  &  - & -  & 7.44 & 77.5 \\
    2 & Classification &   DW  & F & F  &  7.40 & 77.4 \\
    3 & Classification &   BK $L$ & F & F  & 7.40 & 77.8 \\
    4 & Classification &   ParC  & T & F  & 7.40 & 78.0 \\
    5 & Classification &   ParC   & T & T  & 7.41 & \textbf{78.3} \\
    \midrule
    6  & Detection      & Baseline & - & - & 64.98 & 47.20 \\
    7  & Detection      & ParC     & T & F & 64.94 & 47.50 \\
    8  & Detection      & ParC     & T & T & 64.95 & \textbf{47.70} \\
    \midrule
    9  & Segmentation  & Baseline & - & - & 36.30 & 42.17\\
    10 & Segmentation   & ParC     & T & F & 36.27 & 42.23\\
    11 & Segmentation   & ParC     & T & T & 36.28 & \textbf{42.32}\\
    \bottomrule
    \end{tabular}
    \par\vspace{0pt}
    \caption{Ablation study conducted under the framework of ConvNeXt \citep{ConvNext}. DW denotes the split of two one-dimensional (horizontal and vertical) convolutions; BK $L$ denotes the big (one-dimensional) kernel with a length of $L$ (resolution), CC denotes circular convolution and PE denotes position embedding.
    }
    \label{ablation_study_convnext}
\end{table*}

As shown in rows 1-4 of Table~\ref{latency_result}, compared with baseline, ParC-Net is 23\% faster on Rockchip RK3288 and 3.77× faster On DP2000. Besides fewer FLOPs, we believe this speed improvement is also brought by two factors: a) Convolutions are highly optimized by existing tool chains that are widely used to deploy models into these resource-constrained devices; b) Compared with convolutions, transformers are more data bandwidth demanding as computing the attention map involves two large matrices K and Q, whereas in convolutions the kernel is a rather small matrix compared with the input feature map. In case the bandwidth requirement exceeds that of the chip design, the CPU will be left idle waiting for data, resulting in lower CPU utilization and overall slower inference speed.

Results in rows 5-10 show that our proposed ParC-Net universally improves the performances of common ConvNets. For ResNet50 (rows 5-6), it improves accuracy by 0.5 with fewer parameters and FLOPs and almost brings no increase to latency. For MobileNetV2 (rows 7-8) and ConvNeXt (rows 9-10), it improves by 0.9 with a slight increase in budget. 

As is illustrated in Fig.~\ref{fig:parc_tradeoff}, by simply replacing the self-attention or convolution operation with our new operation: 1) our Parc operation improves accuracy of transformer models by 0.2\% on ImageNet classification task and saves as much as 73\% of its inference latency; 2) It improves performance of ConvNets by 0.9\% at most in accuracy on the same task without introducing obvious negative impacts on inference speed and model size. These results demonstrate that ParC, as a plug-and-play operation, can be plugged into different models for different purposes. But whether we apply to ViTs or ConvNets, ParC-based models always have a satisfactory trade-off between performance and complexity.

\subsection{Ablation Study} \label{section:ablation_study}
\textbf{MobileViT Settings.} Using the MobileViT as a baseline model, we further conduct ablation analysis to verify the effectiveness of the 2 components we proposed in our ParC-MetaFormer block. Table~\ref {ablation_study_mobilevit} shows the results.
\begin{enumerate}
    \item Positional Aware Circular Convolution. The proposed ParC has two major characteristics: a) Circular convolution brings a global receptive field; b) PE keeps spatial structure information. Experiment results confirm that both characteristics are important. Results in rows 1-3 show that using a large kernel can also improve accuracy, but the benefit of it reaches a saturation point when kernel size reaches a certain level. These results are consistent with the statement claimed in \citep{ConvNext}. Using ParC can further improve accuracy, as shown in rows 2-3 and 5-6. Introducing PE to ParC is also necessary. As we explained in Section~\ref{subsubsection:parc_operation}, using circular convolution alone can indeed capture global features, but it disturbs the original spatial structures. For the classification task, PE has no impact (rows 5-6). However, for detection and segmentation tasks that are sensitive to spatial location, abandoning PE hurts performances (rows 9-10 and 12-13).
    \item Channel Wise Attention. Results in rows 4 and 5 show that using channel-wise attention can improve performance. But compared with ParC, adopting channel-wise attention brings less benefit to the model, which indicates that the ParC is the main beneficial component in the ParC-MetaFormer block.
\end{enumerate}

\begin{figure*}[t]
    \centering
    \includegraphics[scale=0.38]{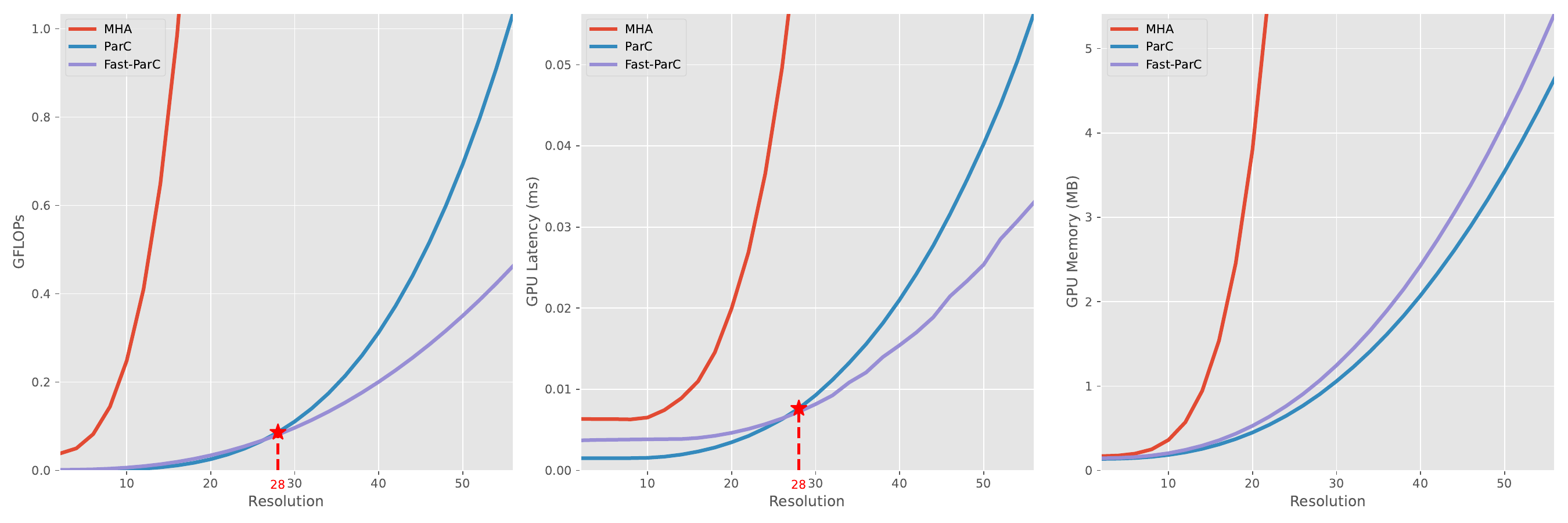}
    \caption{Efficiency comparisons among three types of operations, namely MHA(Multi-Head Attention), ParC, and Fast-ParC. We test the GFLOPs. latency and memory usage with respect to the growth of feature resolution. All data above are collected using an NVIDIA RTX 3090 GPU with batch size 64 and channel number 192. }
\label{fig:fastparc_efficiency}
\end{figure*}

\textbf{ConvNeXt Settings.} For alignment, we conduct the ablation studies on ConvNets using the ConvNeXt-XT as a baseline model. Details are shown in Table~\ref{ablation_study_convnext}. Since the CA (Channel Wise Attention) is not adopted for ConvNets, we study the effectiveness of the following 2 components of ParC. 
\begin{enumerate}
    \item Positional Embedding. Row 4-5, 7-8, and 10-11 show a consistent improvement brought by using positional embedding, which might indicate that positional information is essential for ConvNets, even in the classification task.
    \item Global Circular Convolution. In ConvNeXt-DW-XT, we replace the 7$\times$7 kernel with 1$\times$7 and 7$\times$1 to verify the effectiveness of dimensional-wise convolution. In ConvNeXt-BK-XT, we adopt large vanilla kernels with the same size as stage resolutions (i.e., 13 and 7), but no positional embedding and circular convolution.
    Results in rows 1-2 show that replacing the original two-dimensional 7$\times$7 convolution with dimensional-wise ones (horizontal and vertical) can reduce the model's parameters, but this comes with the cost of accuracy reduction. Rows 2-3 show enlarging the kernel size brings benefits - even by using the DW kernel, it achieves higher accuracy with fewer parameters compared with the baseline. Furthermore, applying ParC to capture the global feature can improve accuracy to a higher level, as shown in rows 2-3 and 4-5. These results indicate that the ParC still owns a significant advantage, which is also consistent with the shortcomings of vanilla convolution that we mentioned in Section~\ref{subsubsection:parc_operation}.
\end{enumerate}

In summary, ParC has two key designs: 1) adopting circular convolution with a global kernel to ensure the global receptive field, and 2) using position embedding to keep position sensitiveness. Both are important.

\subsection{Comparison with the Dilated Convolution}
One conventional approach to enlarge the convolutional receptive field is to use dilated convolution~\citep{dilated_conv} (also known as atrous convolution), which introduces the dilation rate $D$ as an additional parameter to enforce the standard convolution in ConvNets, specifying the spaces between the values in a kernel. By asserting zeros equidistantly into convolutional weights, dilated convolution gets an equivalent large kernel at run-time without parameters increasing.

The ParC/Fast-ParC operation is designed to empower convolution in a more direct and intuitive manner. While dilated convolution gradually expands the receptive field by increasing dilation rates, the ParC/Fast-ParC always ensures the model has a global receptive field. Both of these two variants reduce the parameters/FLOPS complexity increasing caused by large kernels, the former uses zero padding, while the latter uses one-dimensional convolution weight. To compare these two, we conduct comparison experiments on a series of ConvNets. Results are shown in Table~\ref{dilated_classification}. For a fair comparison, we control the dilation rate according to the following equation to ensure the dilated convolution has a global receptive field:
\begin{equation}
    \begin{aligned} \label{dilated_rate}
        & D_{y}=[\frac{H-1}{k_{y}-1}], & P_{y}=[\frac{H-1}{2}] \\
        & D_{x}=[\frac{W-1}{k_{x}-1}], & P_{x}=[\frac{W-1}{2}]
    \end{aligned}
\end{equation}
\begin{table}[t]
    \centering
    \setlength\tabcolsep{3pt}
    \begin{tabular}{llccc}
        \toprule
        Model & Kernel & \makecell{Parms\\(M)} & \makecell{FLOPs\\(G)} & \makecell{Top1 Accu\\(\%)} \\
        \midrule
        \multirow{3}{*}{$\text{ResNet50}^{}$} & Baseline & 25.6  & 4.1 & \multirow{3}{*}{\makecell[l]{79.1\\ 79.1 \\ \textbf{79.6}}} \\
         & Dilated & 25.6 & 4.1 & \\
         & ParC & \textbf{23.7} & \textbf{4.0} & \\
        \midrule
        \multirow{3}{*}{$\text{MobileNetV2}^{}$} & Baseline & 3.51 & 0.6 & \multirow{3}{*}{\makecell[l]{70.2\\ 70.5 \\ \textbf{71.1}}}\\
         & Dilated & 3.51 & 0.6 &  \\
         & ParC & \textit{3.54} & 0.6 &  \\
        \midrule
        \multirow{3}{*}{$\text{ConvNeXt-XT}^{}$} & Baseline & 7.44 & 1.1 & \multirow{3}{*}{\makecell[l]{77.5\\ \textit{77.2} \\ \textbf{78.3}}} \\
         & Dilated & 7.44 & 1.1 & \\
         & ParC & \textbf{7.41} & 1.1 & \\
        \bottomrule
    \end{tabular}
    \caption{Classification results comparison on ImageNet-1K about ParC-based ConvNets and dilated convolution-based ones, both of which try to enlarge the receptive field of networks. The dilated rates are set according to Eq.~\ref{dilated_rate} to ensure global receptive fields. All networks are trained under the ConvNeXt \citep{ConvNext} recipe. }
    \label{dilated_classification}
\end{table}

The experiment results are shown in Table~\ref{dilated_classification}. All ParC-based variants surpass the dilated-convolution-based variants. It demonstrates that ParC is the better operation when trying to enlarge the receptive field since it provides the network with global features of high quality and uses the parameter in a more efficient and effective way.
 
\subsection{Efficiency Comparison among Different Global Operations}
To support the theoretical complexity shown in Table~\ref{tab:fft_theoratical}, we follow up with efficiency comparison experiments among three types of operations, namely MHA(Multi-Head Attention), ParC, and Fast-ParC. We test the GFLOPs, GPU latency, and GPU peak memory usage with respect to the growth of feature resolution using an NVIDIA RTX 3090 GPU with batch size 64 and channel number 192. Results are shown in Fig~\ref{fig:fastparc_efficiency}.

\begin{figure*}[t]
    \centering
    \includegraphics[scale=0.58]{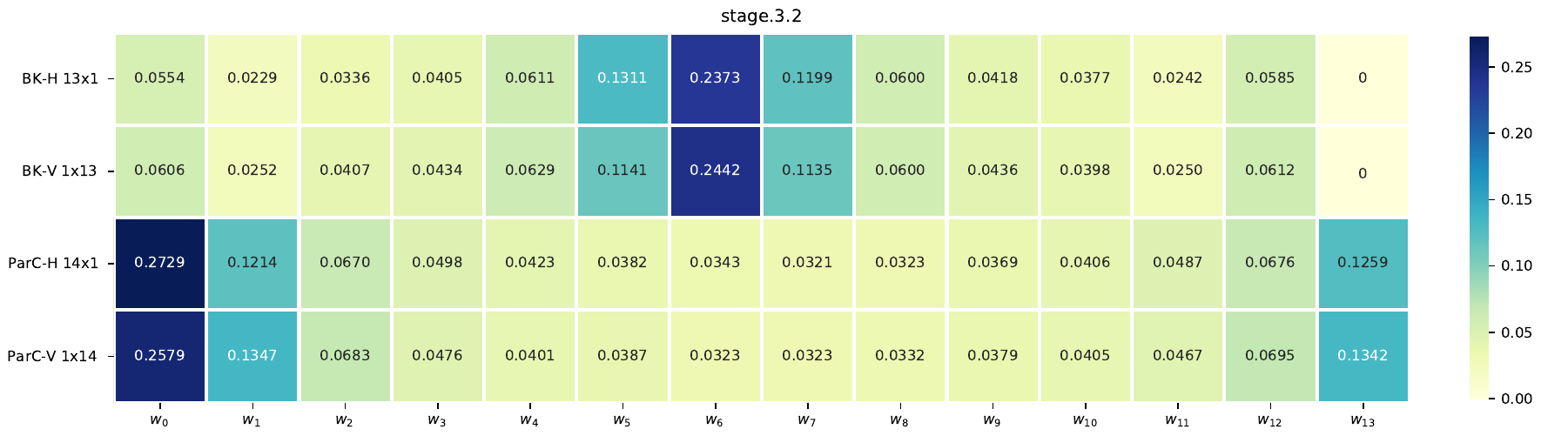}
    \caption{Spatial weight distributions of 1D Big Kernel (BK) and ParC of the 7th block of stage 3, respectively sampled from ConvNeXt-BK-XT and ConvNeXt-ParC-XT. Weight at each spatial location is calculated with the mean absolute value of all channels within one layer. Clearly, weights learned by vanilla big kernel convolution and ParC follow different spatial distributions.}
    \label{fig:parc_weight_pattern}
\end{figure*}

Though all of these three operations are capable of capturing global relationships, their efficiency varies significantly. The GFLOPs, latency, and peak memory usage of MHA  experience a drastic increase between resolutions 10-30, indicating it is impractical to be used in high-feature-resolution tasks. As resolution increases, Fast-ParC emerges as the most efficient regarding FLOPs and latency. The peak memory usage of Fast-ParC, however, is marginally higher than that of the ParC.

A distinctive threshold of ParC/Fast-ParC marked in the figure is $N=28$. When exceeding this resolution, Fast-ParC is generally more efficient, and vice versa. For most ConvNets (e.g., ResNet50, ConvNeXt) with the input image size of 224$\times$224, Fast-ParC is recommended to be applied until the 3rd stage. For downstream tasks with larger input resolutions, Fast-ParC could be introduced in deeper stages. As elaborated before, we can find out the threshold of specific platforms through multiple speed tests and dynamically choose the more efficient method between ParC and Fast-ParC based on resolution, ensuring optimal efficiency.

\section{Discussion}
\subsection{Performance Gap between Classification and Down-stream Tasks}
The MobileViT-ParC shows a similar classification accuracy with MobileViT (+ 0.2\%) but surpasses MobileViT by a clear margin on detection and segmentation. A similar phenomenon appears in experiments based on ConvNets as well, with improvements within + 0.9\% in classification; however, + 2.15\% in semantic segmentation. 

This phenomenon is interesting but explainable, and it demonstrates the concept we included at the beginning of the Experiment part. A good performance in classification cannot guarantee good performance on downstream tasks, which is related to the property of the tasks themselves. Most downstream tasks require more fine-grained spatial information than the classification, which includes two aspects: 1) the ParC operation provides the model with a global receptive field, capturing crucial long-distance dependencies; 2) the positional embedding in the ParC operation strengthens the location information explicitly. As both of these are essential for downstream tasks, the application of ParC benefits the downstream tasks even more significantly than the classification task. Previous research mentions the importance of enlarging effective receptive fields on downstream tasks as well \citep{dilated_conv, large_kernel_matters}.

\subsection{Different Spatial Weights Distribution}
We examined the kernels derived from traditional large kernel convolution and our ParC/Fast-ParC convolution technique. A visualization of these kernels can be seen in Fig.~\ref{fig:parc_weight_pattern}. Interestingly, we observed that while the absolute values of traditional convolution kernel weights typically decrease from the center toward the edges, ParC weights follow the opposite distribution. These two patterns could be aligned via a $K_h/2$-sized shift, which is in accord with the difference between Eq.~\ref{torch_conv} and Eq.~\ref{sptail_parc}. This suggests that, despite its distinct methodology, ParC still inherits the feature extraction rules in traditional convolution. 

\subsection{ParC/Fast-ParC Provides Model with Global Effective Receptive Field}

\begin{figure*}[t]
    \centering
    \includegraphics[scale=0.58]{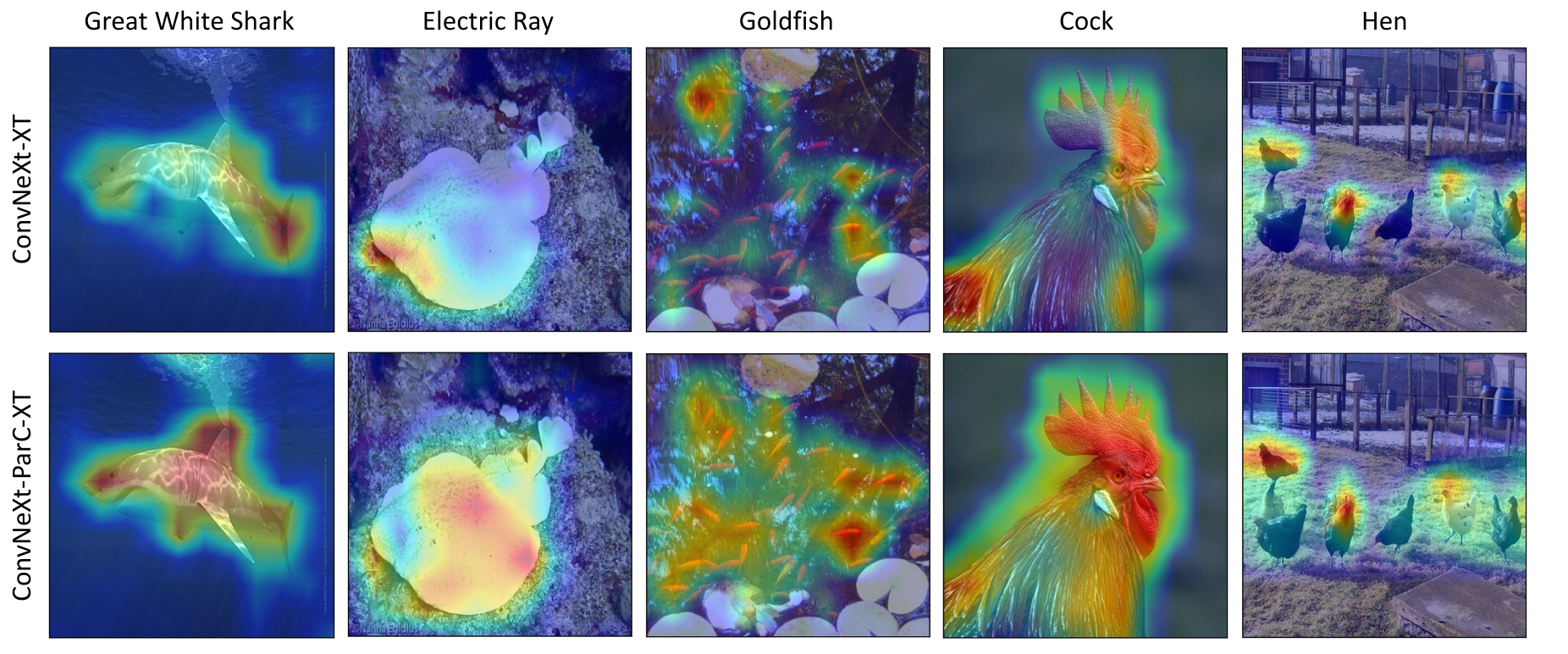}
    \caption{Grad-CAM visualization of vanilla ConvNeXt-XT and ConvNeXt-ParC-XT.}
    \label{fig:grad_cam}
\end{figure*}
Theoretically, a model with ParC/Fast-ParC owns a global receptive field. However, as demonstrated in previous papers~\citep{RF_ERF_PRF}~\citep{ERF2017}~\citep{ERF2021}, the actual receptive field could be saturated due to many concerns. To help understand better how the ParC/Fast-ParC improves models' performance, we test their ERFs ~\citep{ERF2017} and provide the visualization results in this section. 

Based on Fig~\ref{fig:parc_erf_visualization}, we can draw several conclusions:
\begin{enumerate}
    \item \textbf{ParC-ConvNets obtain a global effective receptive field}. Notice models with ParC have a clearly larger ERF than the vanilla ConvNets. Vanilla ConvNets mostly focus on the center round area (red) of pictures, and its sensitivity drops quickly as the radius grows, usually vanishes in the edge/corner region (dark blue). ParC models, however, own a global ERF. Though it focuses especially on the cruciform regions (red), it still has non-zero value on other parts (yellow or shallow blue). A more obvious merit is that ParC-ConvNeXt's ERF does not decay radially. 
    \item \textbf{ERF from different architectures show significantly different shapes}. The ERF of ResNet50 exhibits a \textit{square} outline; the ConvNext-XT and MobileNetV2, however, possess a much more round outline. This might be closely related to the activation function the networks used (i.e., ReLU, ReLU6, and GELU). A similar phenomenon~\citep{ERF2017} is reported to exist between Sigmoid and ReLU, too.  Apart from the difference in outline, ConvNeXt's ERF also shows a square shadow inside of its round outline. This might be the result of its non-overlap down-sampling layer. In contrast, ResNet and MobileNetV2 have a round shadow in their round outline.
    \item \textbf{The ERF of ConvNeXt is relatively larger than ResNet50 and MobileNetV2}. This is a reasonable conclusion - since ConvNeXt uses a 7$\times$7 kernel to replace a 3$\times$3 kernel in ResNet50. However, this only expands local receptive fields. The figure shows that in vanilla ConvNeXt-XT's ERF, the magnitude still decays radially. ParC integrates global information, offering ConvNeXt-ParC-XT an effective receptive field over the whole image. As a consequence, ConvNeXt-ParC-XT obtains a better performance, indicating that vanilla large kernels are not as good as global kernels. A parallel work, RepLKNet~\citep{RepKLNet}, also mentioned that increasing the kernel size of ConvNeXt further to 31$\times$31 could be beneficial.
\end{enumerate}

\subsection{ParC Helps Model to Form More Comprehensive Focuses}
The proposed ParC-based models generally follow this rule: extract local information in its shallow layer, then integrate global information in this deeper layer. This helps the model to focus on semantic information that might be hard for vanilla models to catch. We utilize Grad-CAM \citep{GradCAM} to visualize the semantic important regions of vanilla ConvNets and ParC-ConvNets. From Fig.~\ref{fig:grad_cam}, the benefit brought by ParC in focusing on semantic important regions could be concluded into two major points: 
\begin{enumerate}
    \item \textbf{For a picture with a single instance, ParC helps the model capture the whole instance instead of focusing on some specific parts of it.} The images in columns 1, 2, and 4 could support this statement. While capturing features, Vanilla ConvNet focuses on distinguishing features like shark tails, cockscomb, or the periphery of instances. But ParC-ConvNet focuses on the entire instance, while those distinguishing features are only sub-regions within.
    \item \textbf{For pictures with multiple instances, ParC helps the model capture more individuals.} As shown in columns 3 and 5 in Fig~\ref{fig:grad_cam}, when facing a picture with multiple instances, vanilla ConvNet tends to miss some of them, but ParC-ConvNet could capture all. This characteristic of ParC might be beneficial for dense prediction tasks.
\end{enumerate}

\section{Conclusion}
We designed a novel plug-and-play operation named ParC (Positional Aware Circular Convolution). ParC owns a global receptive field like self-attention used in ViT but could be supported more conveniently by different hardware platforms since it uses pure convolution operation. We demonstrate it boosts networks' performance on classification whether to be inserted on transformer-based networks or convolution-based networks. Besides, these ParC-based models show superiority on downstream tasks as well. We also analyzed the inner mechanism and its difference compared with vanilla big kernel convolution and then gave some convincing explanations for its superiority. Fast-ParC, an FFT-based version of ParC, is also proposed for applying ParC in conditions of large resolution. Fast-ParC operation is capable of maintaining a low computation budget even with high input resolution, making ParC a competitive general choice for most computer vision tasks.

\section{Data Availability Statement}
In this paper, we conducted experiments on four datasets: ImageNet-1K~\citep{ImageNet}, MS-COCO~\citep{MS_COCO}, PASCAL VOC~\citep{VOC}, and ADE20K~\citep{ADE20K}. All these datasets are publicly available. Readers can refer to the respective citations to find the download links.

\bibliographystyle{spbasic}      
\bibliography{ref}


\end{document}